\documentclass{article}

\usepackage{microtype}
\usepackage{graphicx}
\usepackage{subcaption}
\usepackage{booktabs} 
\usepackage{amsmath}    
\usepackage{bm}         
\usepackage{hyperref}
\usepackage{placeins}
\usepackage{float}

\usepackage[accepted]{icml2026}

\usepackage{amsmath}
\usepackage{amssymb}
\usepackage{mathtools}
\usepackage{amsthm}
\usepackage{makecell}

\usepackage[capitalize,noabbrev]{cleveref}

\theoremstyle{plain}

\theoremstyle{definition}

\theoremstyle{remark}

\usepackage[textsize=tiny]{todonotes}
\usepackage{multirow}
\icmltitlerunning{PIS: A Generalized Physical Inversion Solver for Arbitrary Sparse Observations via Set Conditioned Flow Matching}

\begin{document}

\twocolumn[
  \icmltitle{PIS: A Generalized Physical Inversion Solver for Arbitrary Sparse Observations via Set Conditioned Flow Matching}



  \icmlsetsymbol{equal}{*}

\begin{icmlauthorlist}
\icmlauthor{Weijie Yang}{ucb}
\icmlauthor{Xun Zhang}{tongji}
\icmlauthor{Simin Jiang}{tongji}
\icmlauthor{Yubao Zhou}{imsia}
\end{icmlauthorlist}

\icmlaffiliation{ucb}{School of Information, University of California, Berkeley, USA}
\icmlaffiliation{tongji}{College of Civil Engineering, Tongji University, Shanghai, China}
\icmlaffiliation{imsia}{Institute of Mechanical Sciences and Industrial Applications (IMSIA), Ecole Nationale Supérieure de Techniques Avancées, France}

\icmlcorrespondingauthor{Simin Jiang}{jiangsimin@tongji.edu.cn}

\icmlkeywords{Physics-Informed Flow Matching, Geophysics, Inverse Problems, Machine Learning}

\vskip 0.3in
]



\printAffiliationsAndNotice{}  

\begin{abstract}
The estimation of high-dimensional physical parameters constrained by partial differential equations (PDEs) from limited and indirect measurements is a highly ill-posed problem. Traditional methods face significant accuracy and efficiency bottlenecks, particularly when observations are sparse, irregularly sampled, and constrained by real-world sensor placement. We propose the Physical Inversion Solver (PIS), a unified framework that couples Set-Conditioned Flow Matching with a Cosine-Annealed Sparsity Curriculum (CASC) to enable stable inversion from arbitrary, off-grid sensors even under minimal guidance. By leveraging straight-path transport, PIS achieves instantaneous inference (50 NFEs), offering orders-of-magnitude speedup over iterative baselines. Extensive experiments demonstrate that PIS reduces error by up to 88.7\% under extreme sparsity ($<1\%$) across subsurface characterization, wave-based characterization, and structural health monitoring, while providing robust uncertainty quantification for optimal sensor placement.
\end{abstract}

\section{Introduction}
\label{sec:intro}

Inverse problems governed by partial differential equations (PDEs) are fundamental to scientific discovery, aiming to infer high-dimensional parameters from limited system responses. Whether characterizing subsurface hydraulic conductivity \cite{zhang2024integration}, identifying structural damage in infrastructure \cite{ni2021deep}, or inverting wave velocity models \cite{pan2025transcending}, these tasks share a core challenge: observations are inherently sparse, noisy, and unstructured, rendering the inverse mapping strictly ill-posed.

Despite the promise of deep learning, current paradigms face a fundamental trilemma between flexibility (handling arbitrary sensors), efficiency (fast inference), and stability (convergence). While operator learners (e.g., FNO) \cite{li2021fourier} and diffusion models \cite{song2020score} have made strides, they either struggle with off-grid and arbitrary observations or suffer from slow stochastic sampling.

To resolve this trilemma, we propose the Physical Inversion Solver (PIS), a set-conditioned flow matching framework. PIS unifies the processing of arbitrary, off-grid sensors via permutation-invariant embeddings and accelerates inference via deterministic probability flows. To prevent mode collapse under minimal guidance, we further introduce a Cosine-Annealed Sparsity Curriculum (CASC). Our main contributions are:

\begin{enumerate}
    \item \textbf{Robust Generalization under Sparsity:} We propose the Set-Conditioned Transformer U-Net (SCTU-Net) trained via CASC. This synergy allows PIS to process unstructured, off-grid inputs and prevents posterior collapse under extreme sparsity, enabling zero-shot generalization across variable sensor layouts.
    
    \item \textbf{Real-Time Efficiency:} Leveraging the straight-path property of Flow Matching, PIS achieves high-fidelity inversion within just 50 NFEs. This orders-of-magnitude speedup enables instantaneous diagnostics for time-sensitive engineering tasks.
    
    \item \textbf{Information-Theoretic Insight:} We quantify the fundamental information limits via Shannon Entropy, providing data-driven guidelines for optimal sensor placement in engineering domains with real constraints.
\end{enumerate}

\section{Related Work}
\label{sec:related}
We summarize machine learning paradigms for physical inversion below, with an extended review in Appendix~A.

While traditional Bayesian inference such as Ensemble Kalman Filter (EnKF) and Markov Chain Monte Carlo (MCMC) offers theoretical soundness, it suffers from severe computational bottlenecks due to thousands of simulations \cite{emerick2013ensemble,wang2024physics}. To overcome this, paradigms have shifted to deep learning. However, Neural Operators \cite{li2021fourier}, while efficient, face two hurdles: they are deterministic and grid-dependent.

To restore probabilistic rigor, Diffusion Models \cite{song2020score} function as learned Bayesian priors but suffer from slow convergence. Flow Matching (FM) \cite{lipman2022flow} resolves this efficiency bottleneck via straight-path ODE integration. Yet, standard FM lacks the geometric inductive bias to process the unstructured, permutation-invariant sensor sets inherent to physics.

PIS bridges these limitations by integrating the efficiency of FM with a set-native architecture. This positions PIS as a robust solver for the Bayesian Inverse formulation, defined formally in the next section.

\section{Preliminaries: Theoretical Foundations}

\subsection{Problem Setting: Physical Inversion}
Physical inversion refers to the problem of estimating unobservable physical parameters from indirect measurements, governed by known physical processes.

\begin{equation}
    \mathbf{y} = \mathcal{G}(\mathbf{x}) + \bm{\epsilon}, \quad \mathbf{x} \in \mathcal{X}
\end{equation}

Here, $\mathcal{G}: \mathbb{R}^\mathbf{N_x} \mapsto \mathbb{R}^\mathbf{N_y}$ represents the forward operator, mapping physical parameters to observed data, governed by physical laws such as porous-media fluid dynamics, wave propagation, or structural mechanics. $\mathbf{N_x}$ is the dimension of the parameter vector $\mathbf{x}$ and $\mathbf{N_y}$ is the dimension of the observation vector $\mathbf{y}$; $\mathbf{x}$ refers to the generalized physical parameters; $\mathbf{y}$ denotes the observed data; and $\bm{\epsilon} \sim \mathcal{N}(\mathbf{0}, \bm{\Sigma})$ denotes the observation error vector, which is assumed to follow a Gaussian distribution.

In such systems, the measurements $\mathbf{y}$ represent the observable effects, while the physical parameter $\mathbf{x}$ is the cause that generated them. Thus, inverse modeling aims to estimate $\mathbf{x}$ from observations by learning (or constructing) an inverse operator that maps measured data back to the underlying physical parameters, i.e.,
\begin{equation}
    \mathbf{x} = \mathcal{G}^{-1}(\mathbf{y})
\end{equation}
whenever such an inverse exists. However, classical and geophysical inverse problems are typically ill-posed in the sense of Hadamard \cite{hadamard1923}. Specifically, the forward operator $\mathcal{G}$ is often non-injective (leading to non-uniqueness), unstable under small perturbations in data, or defined over high or infinite dimensional parameter spaces \cite{kaipio2005,stuart2010inverse,tarantola1987inverse}. As a result, the inverse mapping $\mathcal{G}^{-1}$ may fail to exist, may not be unique, or may lack continuous dependence on the data --- violating the conditions of well-posedness.

\subsection{Bayesian Inversion}
A natural approach to addressing ill-posedness and quantifying uncertainty in physical inversion problems is to formulate them in a Bayesian framework \cite{tarantola1987inverse,kaipio2005,stuart2010inverse}. Given sparse and irregular observations $\mathbf{y}_{\omega} = \{\mathbf{y}(\mathbf{x}_i)\}_{\mathbf{x}_i \in \omega}$ and a forward operator $\mathcal{G}$, the posterior over the unobservable physical quantity $\mathbf{x}$ is
\begin{equation}
    p(\mathbf{x} \mid \mathbf{y}_{\omega}) \propto p(\mathbf{y}_{\omega} \mid \mathbf{x}) p(\mathbf{x})
\end{equation}
where $p(\mathbf{x})$ is the prior over physical fields, and $p(\mathbf{y}_{\omega} \mid \mathbf{x})$ is the likelihood induced by the forward model restricted to observation locations $\omega$. In the sparse data regime, the likelihood is weakly informative, as only a small fraction of the output of $\mathcal{G}(\mathbf{x})$ is constrained by observations. This results in posterior broadening, increased multi-modality, and significant non-uniqueness \cite{zhao2024bvi}.

Explicit Bayesian solvers (e.g., MCMC, EnKF) offer a principled framework for uncertainty quantification but face severe computational bottlenecks in high-dimensional settings, as each posterior evaluation requires repeated, expensive forward or adjoint solves \cite{kaipio2005,stuart2010inverse}. While acceleration strategies like surrogates or dimensionality reduction exist \cite{mo2020integration,2014Likelihood}, they generally lack the flexibility to natively ingest sparse, off-grid observations. This limitation necessitates the development of robust inference methods that can directly map arbitrary sensor inputs to the posterior without iterative PDE solving.

\subsection{Information-Theoretic Sparsity Limits}
The presence of sparse and irregular observation patterns substantially exacerbates the difficulty of Bayesian physical inversion. Observations carry information about the underlying physical field, and quantifying how much they constrain the posterior is critical for both inference and experimental design. This motivates an information-theoretic formulation, where mutual information provides a principled metric for assessing how observational sparsity affects posterior uncertainty.

\begin{equation}
    I(\mathbf{x}; \mathbf{y}_{\omega}) = H(\mathbf{x}) - H(\mathbf{x} \mid \mathbf{y}_{\omega})
\end{equation}

where $H(\cdot)$ denotes the Shannon entropy \cite{shannon1948,cover2006}. As $|\omega|$ decreases, the entropy reduction $H(\mathbf{x}) - H(\mathbf{x} \mid \mathbf{y}_{\omega})$ rapidly vanishes---often approaching zero in high-dimensional systems. This phenomenon---referred to as information collapse \cite{lindley1956,alexanderian2014}---indicates that sparse observations fail to sufficiently constrain the posterior, preventing stable contraction. We employ this metric to evaluate the sparsity extreme under different science scenarios in the Experiments section.

\section{PIS: Physical Inversion Solver}

\begin{figure*}[t]  
    \centering
    \includegraphics[width=\textwidth]{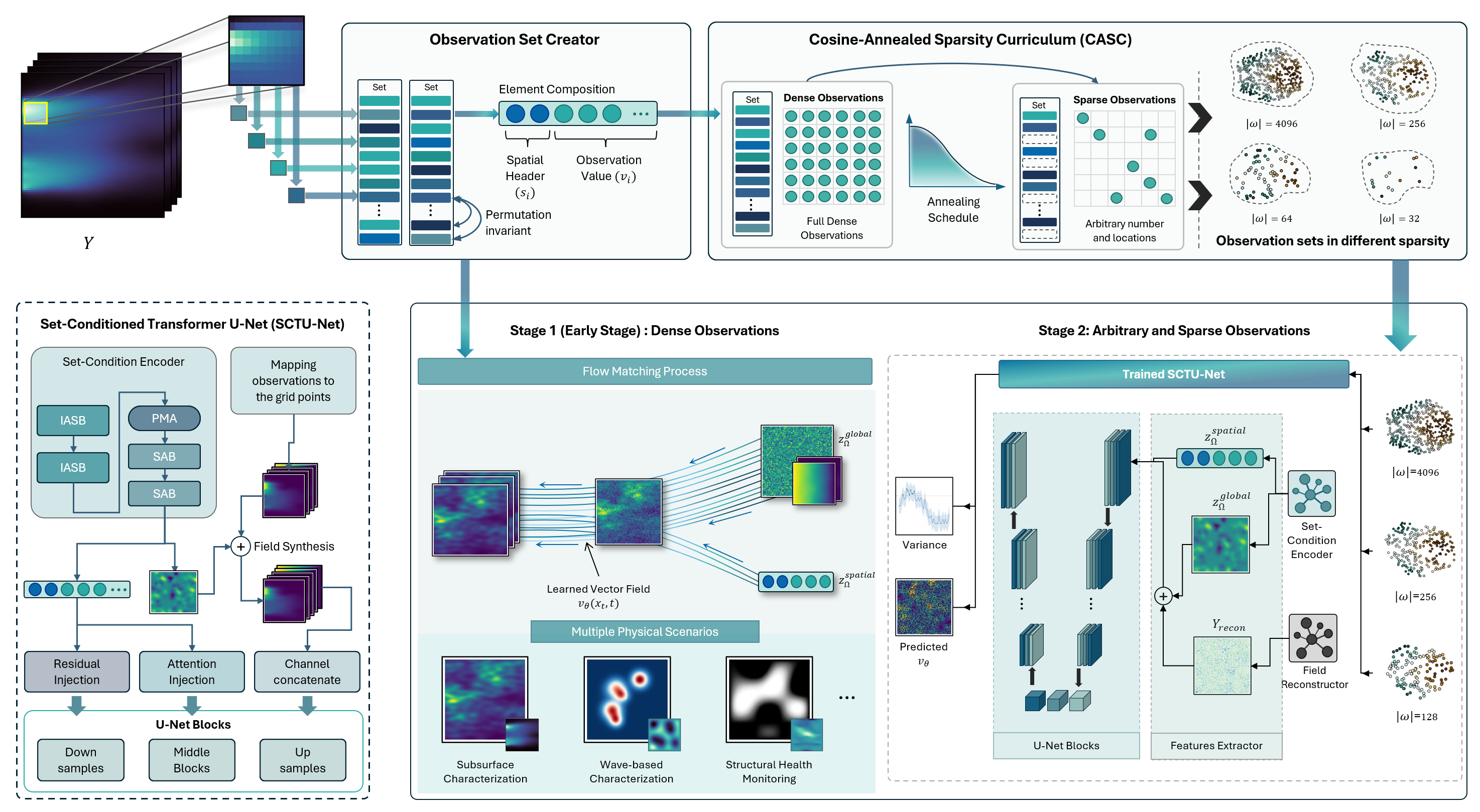}
    \caption{\textbf{PIS Model architecture}. Sparse measurements are represented as permutation-invariant sets and randomly sampled using a cosine-annealed schedule to simulate arbitrariness. The model first uses dense observations to learn global field structure, and then adapts to randomly sparse observations. A Set-Conditioned U-Net injects set-level conditioning into the Flow Matching process, enabling robust inversion from irregular and variable-sized observation sets by predicting $v_\theta$ with variance regularization.}
    \label{fig:c3_architecture}
\end{figure*}

\subsection{Conditional Flow Matching for Inversion}
Building upon the Bayesian framework defined in Section~2.2, our goal is to characterize the posterior distribution $p(\mathbf{x} \mid \mathbf{y}_{\omega})$. Traditionally, generating samples from this posterior using diffusion models involves complex stochastic differential equations (SDEs) and iterative de-noising \cite{song2021score}. In this work, we transition to the Conditional Flow Matching (CFM) paradigm to construct a deterministic probability path between a simple noise distribution and the complex physical posterior.

\textbf{Probability Paths and Vector Fields.} Let $p_1(\mathbf{x})$ be the target posterior distribution $p(\mathbf{x} \mid \mathbf{y}_{\omega})$ and $p_0(\mathbf{x}_0)$ be a standard Gaussian prior $\mathcal{N}(\mathbf{0}, \mathbf{I})$. We define a time-dependent probability path $p_t(\mathbf{x})$ that couples $p_0$ and $p_1$ through a linear interpolation:
\begin{equation}
    \mathbf{x}_t = \psi_t(\mathbf{x}_1) = (1-t) \mathbf{x}_0 + t \mathbf{x}_1, \quad t \in [0,1]
\end{equation}
where $\mathbf{x}_1 \sim p_1(\mathbf{x})$ and $\mathbf{x}_0 \sim p_0(\mathbf{x})$. This formulation ensures that each noise sample $\mathbf{x}_0$ is transported to a physical parameter $\mathbf{x}_1$ along a straight-line trajectory. The ideal velocity vector field $u_t(\mathbf{x} \mid \mathbf{x}_1)$ generating this path is defined as the time derivative of $\psi_t$:
\begin{equation}
    u_t(\mathbf{x} \mid \mathbf{x}_1) = \frac{d}{dt} \psi_t(\mathbf{x}_1) = \mathbf{x}_1 - \mathbf{x}_0.
\end{equation}
This straight-path property is a significant advantage over diffusion approaches, as it minimizes the curvature of the generation trajectory, thereby enabling high-fidelity inversion with fewer function evaluations (NFEs) \cite{liu2023flow}.

\textbf{CFM Objective.} Since the ground-truth vector field $u_t$ is inaccessible, we parameterize a neural network $v_{\theta}(\mathbf{x}_t, t, \mathbf{y}_{\omega})$ to approximate the velocity field. Given the set-conditioned observations $\mathbf{y}_{\omega}$ defined in Section~2.2, the training objective is to minimize the conditional flow matching loss:
\begin{equation}
\mathcal{L}_{\text{CFM}}(\theta) = \mathbb{E}_{t, \mathbf{x}_1, \mathbf{x}_0} \left[ \| v_{\theta}(\mathbf{x}_t, t, \mathbf{y}_{\omega}) - (\mathbf{x}_1 - \mathbf{x}_0) \|^2 \right]
\end{equation}
Unlike score-based methods that estimate the score function $\nabla \log p_t(\mathbf{x})$, PIS directly regresses the displacement field. This approach is more numerically stable in the highly under-determined regimes discussed in Section~2.3, as it avoids the singularities often encountered in score estimation at small $t$ \cite{albergo2023stochastic}.

\subsection{Backbone: Set-Conditioned Transformer U-Net}
\label{subsec:architecture}

We propose the SCTU-Net to process unstructured sensors without heuristic interpolation. The architecture features a dual-path encoder and a conditioned U-Net decoder.

\textbf{Dual-Path Set Encoding.}
A Set Transformer with Induced Set Attention Blocks (ISAB) encodes observations $\mathbf{y}_{\omega}$ into latent features $\mathbf{H}^{(L)}$. The encoder bifurcates into:
\begin{itemize}
    \item \textbf{Global Context} ($\mathbf{F}_{\text{global}}$): $\mathbf{H}^{(L)}$ is aggregated via PMA. To preserve information flow, the final descriptor $\mathbf{F}_{\text{global}}$ is constructed by concatenating the attention-pooled features with the residual identity.
    \item \textbf{Spatial Map} ($\mathbf{F}_{\text{spatial}}$): Learnable queries $\mathbf{Q}_{\text{grid}} \in \mathbb{R}^{H \times W \times d}$ (initialized as positional embeddings) interrogate $\mathbf{H}^{(L)}$ via cross-attention to produce the grid-aligned map $\mathbf{F}_{\text{spatial}}$.
\end{itemize}

\textbf{Conditioning Integration.}
The derived set embeddings are injected into the U-Net backbone $v_{\theta}$ via two mechanisms:
(i) Field Synthesis: We first construct a raw sparse tensor $\mathbf{M}_{\text{obs}}$ by mapping observations to the nearest grid points. The U-Net input is then formed by concatenating the noisy state $\mathbf{x}_t$, the learned spatial map $\mathbf{F}_{\text{spatial}}$, and the explicit cues $\mathbf{M}_{\text{obs}}$.
(ii) Global Integration: The macroscopic descriptor $\mathbf{F}_{\text{global}}$ modulates the feature maps via Adaptive Group Normalization.

\subsection{Cosine-Annealed Sparsity Curriculum}
\label{subsec:training_strategy}

Directly regressing conditional vector fields under extreme sparsity ($<1\%$) leads to optimization instability due to weak guidance signals. To circumvent this, we introduce the Cosine-Annealed Sparsity Curriculum (CASC), which stabilizes convergence by progressively increasing the ill-posedness of the inversion task.

\textbf{Sparsity Annealing.}
We dynamically anneal the observation cardinality $|\omega|_k$ from a dense regime ($|\omega|_{\max}$) to the target sparsity ($|\omega|_{\min}$) over training iterations $k$. The schedule follows a cosine trajectory:

\begin{equation}
|\omega|_k = |\omega|_{\min} + \frac{1}{2} \left( |\omega|_{\max} - |\omega|_{\min} \right) \left( 1 + \cos\left(\frac{k\pi}{K}\right) \right)
\end{equation}

Theoretically, this schedule functions as a continuation method on the optimization landscape. In the early dense regime ($|\omega| \approx |\omega|_{\max}$), the problem is well-posed, effectively serving as a convex relaxation that guides the model to a valid basin of attraction.
From an information-theoretic perspective, this dense initialization ensures high mutual information $I(\mathbf{x}; \mathbf{y}_\omega)$, preventing the posterior collapse often observed when $I(\mathbf{x}; \mathbf{y}_\omega) \to 0$. As sparsity increases, the curriculum continuously initializes the difficult optimization steps with the trajectory learned from the high-information regime, ensuring robust convergence.

\textbf{Hybrid Optimization Objective.}
To enforce robustness against observational stochasticity, we employ a consistency-regularized objective:

\begin{equation}
\mathcal{L}_{\text{total}} = \mathbb{E}_{t,\mathbf{x}_1} \left[ \| v_t - u_t \|^2 + \lambda \| \mathbf{\Sigma}_t \|^2 \right]
\end{equation}

Crucially, $\mathbf{\Sigma}_t$ represents the intra-sample variance, computed across $K$ stochastic observation masks $\{\mathbf{y}_{\omega}^{(k)}\}_{k=1}^K$ generated for the same physical state $\mathbf{x}_1$. Unlike batch variance, minimizing $\mathbf{\Sigma}_t$ explicitly penalizes sensitivity to sensor placement without suppressing the physical diversity across independent samples.

\subsection{Inference via ODE Integration}
Once trained, PIS performs inversion by solving the Initial Value Problem of the learned probability flow ODE. Starting from noise $\mathbf{x}_0 \sim \mathcal{N}(\mathbf{0}, \mathbf{I})$, the trajectory to the posterior sample $\mathbf{x}_1$ is governed by:

\begin{equation}
d\mathbf{x}_t = v_{\theta}(\mathbf{x}_t, t, \mathbf{y}_{\omega}) dt, \quad \mathbf{x}_0 \sim \mathcal{N}(\mathbf{0}, \mathbf{I})
\end{equation}

where $\mathbf{y}_{\omega}$ denotes the frozen set-conditioned features that steer the flow toward physically consistent manifolds.

\textbf{Efficiency via Straight Path.}
A critical advantage of our optimal transport objective is the induction of straight-path trajectories with minimal curvature. Unlike diffusion models requiring traversing complex stochastic paths, PIS approximates a constant-velocity flow. This geometric regularity allows standard solvers such as Euler to achieve high-fidelity inversion in only 50 NFEs, enabling real-time inference for engineering tasks.

\textbf{Uncertainty Quantification.}
Although the ODE is deterministic for a given $\mathbf{x}_0$, PIS naturally quantifies uncertainty by integrating a batch of independent noise samples. The resulting ensemble $\{\mathbf{x}_0^{(k)}\}_{k=1}^K$ captures the multimodal posterior distribution, allowing for the direct computation of pixel-wise variance and confidence intervals to identify regions of information collapse.

\section{Experiments}
\textbf{Benchmarks.}
We evaluate PIS on three PDE systems: subsurface characterization under Darcy flow and advection-diffusion equations, wave-based characterization under Helmholtz equations and structural health monitoring (SHM). Comparisons are made against three paradigms: (i) Neural Operators (FNO \cite{li2021fourier}, NIO \cite{molinaro2023neural}); (ii) Physics-Informed Optimization (PINN \cite{raissi2019pinn}); and (iii) Standard Flow Matching (NFM) \cite{lipman2022flow}. 
To ensure a fair and rigorous comparison, sparse, off-grid inputs for all baselines are pre-processed via linear interpolation to construct dense grid representations. This strategy provides a significantly stronger initialization than naive zero-filling. Additionally, a diffusion-based variant (PIS-DDPM) is analyzed in Appendix~F to validate framework generality.

\textbf{Protocol.}
We simulate unstructured, off-grid sensors by randomly sampling observation sets with cardinalities $|\omega|$ ranging from 12 to 64. To rigorously quantify performance, we report RMSE and SSIM averaged over 200 independent realizations to account for stochastic sensor placement. Inference utilizes a deterministic Euler solver with a fixed computational budget of NFE $= 50$.

\subsection{Physical Scenarios and Datasets}
\textbf{Dataset Specifications.} 
Across all benchmarks, we utilize datasets of 5,000 samples defined on a $64 \times 64$ discretized domain. The target is a scalar physical parameter field $\mathbf{x} \in \mathbb{R}^{1 \times 64 \times 64}$. To handle unstructured sparsity, inputs are formatted as unordered sets $\mathbf{y}_{\omega} = \{(\mathbf{s}_i, \mathbf{v}_i)\}_{i=1}^K$ with varying cardinality $K$. Each element concatenates the 2D spatial coordinate $\mathbf{s}_i \in \mathbb{R}^2$ with the observed observation values $\mathbf{v}_i \in \mathbb{R}^V$, resulting in a per-point feature dimension of $2 + V$. Further details are provided in Appendix~C.

\textbf{Subsurface Characterization.}
We estimate the heterogeneous hydraulic conductivity field $K(\mathbf{x})$ from sparse measurements of hydraulic head and solute concentration. The system is governed by steady-state Darcy flow coupled with advection-dispersion equations. Log-conductivity fields are generated via Karhunen--Loève Expansion (KLE) to enforce realistic spatial correlations, with forward simulations performed using MODFLOW \cite{MODFLOW} and MT3DMS \cite{MT3DMS}.

\textit{Practical Significance:} Accurate conductivity mapping is critical for groundwater resource management and contaminant remediation. In real-world hydrogeology, subsurface data is inherently sparse because drilling observation wells is destructive and expensive \cite{carrera2005inverse}.

\textbf{Wave-based Characterization.}
We utilize the public Helmholtz benchmark (camlab-ethz) to inverse spatially varying wavenumbers from partial wavefield observations. The forward model follows the 2D Helmholtz equation $\nabla^2 u + k^2 u = f$, representing steady-state wave propagation typical in seismic imaging. Inputs consist of complex-valued wavefield intensities sparsely sampled from the full resolution domain.

\textit{Practical Significance:} This task mirrors challenges in geophysical exploration. Physical constraints typically restrict receivers (e.g., geophones) to surface arrays or limited boreholes, making the recovery of internal medium properties from sparse wavefield samples a fundamental yet ill-posed challenge \cite{virieux2009overview}.

\textbf{Structural Health Monitoring.}
We address a static inverse elasticity problem for Non-Destructive Evaluation (NDE). The objective is to reconstruct the Young's modulus field $E(\mathbf{x})$ of a two-phase heterogeneous medium (matrix with random stiff inclusions) solely from sparse displacement measurements. Equilibrium displacement fields are generated via Finite Element Method (FEM) under plane strain conditions with uniform compression.

\textit{Practical Significance:} This scenario is vital for civil and aerospace infrastructure safety, identifying internal stiffness degradation is a key engineering objective. However, deploying dense sensor arrays is economically and logistically infeasible, necessitating inversion from limited, discrete sensor locations \cite{ostachowicz2019optimization}.

\begin{figure*}[t]  
    \centering
    \includegraphics[width=\textwidth]{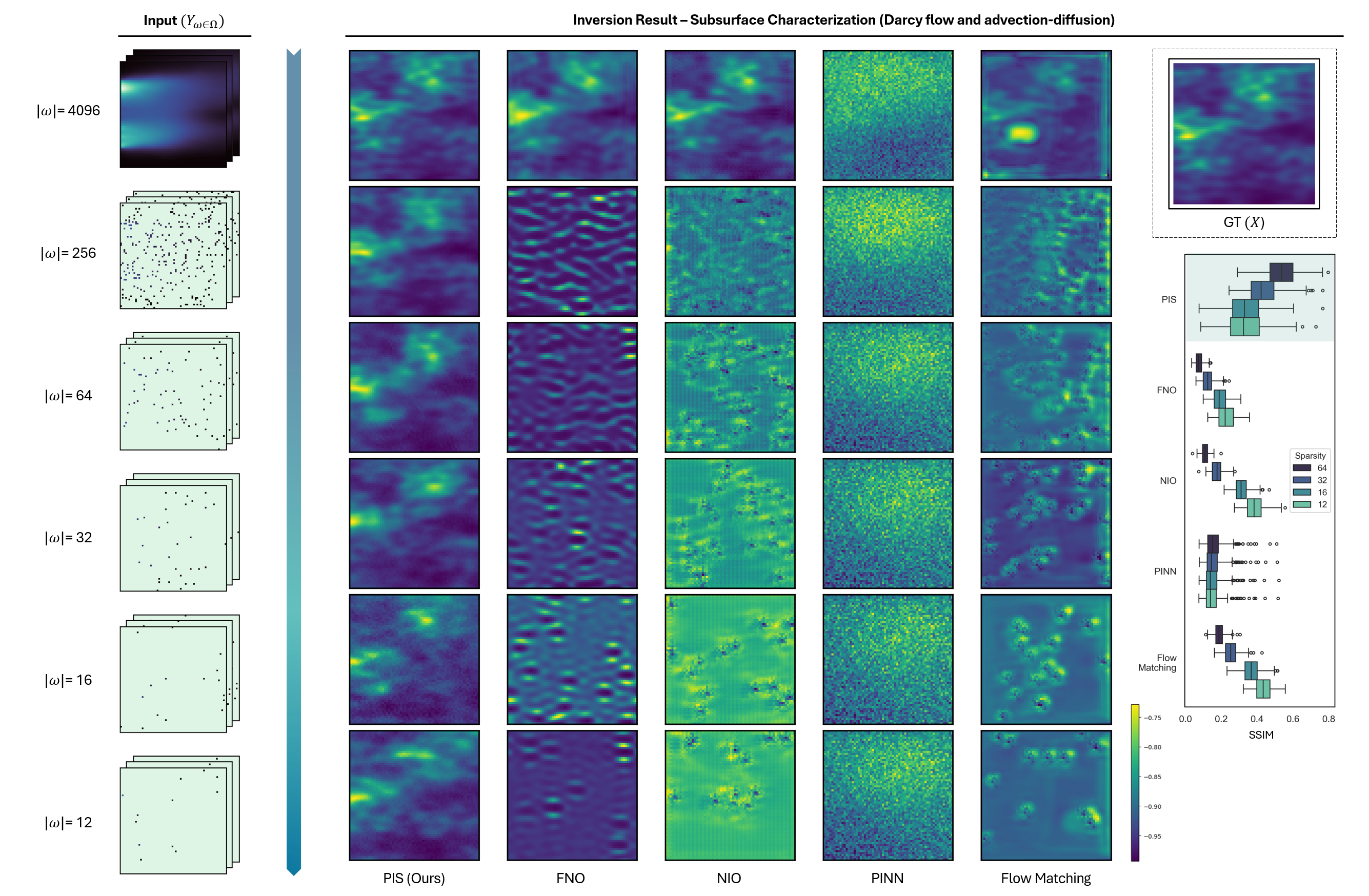}
    \caption{\textbf{Inversion results for subsurface characterization under varying sparsity.} Rows correspond to decreasing observation budgets $|\omega| \in \{4096, 256, 64, 32, 16, 12\}$. While baselines exhibit significant artifacts or over-smoothing at extreme sparsity ($|\omega| = 12$), PIS maintains structural fidelity, accurately resolving high-permeability channels. Right panel: SSIM distributions confirm PIS consistently outperforms state-of-the-art solvers (FNO, NIO, PINN, NFM).}
    \label{fig:3_inversion_result}
\end{figure*}

\subsection{Inversion Quality and Efficiency Analysis}

Ensuring physically consistent inversion under extreme sparsity ($<1\%$) remains a challenge where classical solvers often collapse. Following established benchmarks \cite{jin2017deep}, this section evaluates whether PIS can preserve both pixel-wise accuracy and topological fidelity in this severely ill-posed regime, specifically focusing on the trade-off between perceptual plausibility and physical correctness.

\begin{table*}[t]
\centering
\small
\setlength{\tabcolsep}{3pt}
\caption{Inversion quality in terms of RMSE (lower is better) and SSIM (higher is better) across different observation budgets.}
\label{tab:inv_quality}
\resizebox{\textwidth}{!}{%
\begin{tabular}{ll*{6}{cc}}
\toprule
\multirow{2}{*}{Dataset} & \multirow{2}{*}{Model}
& \multicolumn{2}{c}{Dense: 4096}
& \multicolumn{2}{c}{Sparse: 256}
& \multicolumn{2}{c}{Sparse: 64}
& \multicolumn{2}{c}{Sparse: 32}
& \multicolumn{2}{c}{Sparse: 16}
& \multicolumn{2}{c}{Sparse: 12} \\
\cmidrule(lr){3-4}\cmidrule(lr){5-6}\cmidrule(lr){7-8}
\cmidrule(lr){9-10}\cmidrule(lr){11-12}\cmidrule(lr){13-14}
& & RMSE & SSIM & RMSE & SSIM & RMSE & SSIM & RMSE & SSIM & RMSE & SSIM & RMSE & SSIM \\
\midrule
\multirow{5}{*}{\makecell[l]{Subsurface\\Characterization\\(Darcy flow and advection-diffusion)}}
& PINN
& $0.045\pm0.021$ & $0.249\pm0.092$
& $0.164\pm0.019$ & $0.172\pm0.068$
& $0.160\pm0.019$ & $0.167\pm0.067$
& $0.159\pm0.019$ & $0.162\pm0.066$
& $0.158\pm0.019$ & $0.159\pm0.066$
& $0.158\pm0.019$ & $0.158\pm0.066$ \\

& FNO
& $0.020\pm0.010$ & $0.836\pm0.039$
& $1.782\pm0.125$ & $0.026\pm0.012$
& $0.963\pm0.057$ & $0.076\pm0.023$
& $0.780\pm0.040$ & $0.126\pm0.035$
& $0.669\pm0.030$ & $0.195\pm0.044$
& $0.643\pm0.029$ & $0.229\pm0.050$ \\

& NIO
& $0.024\pm0.019$ & $0.872\pm0.068$
& $0.449\pm0.036$ & $0.087\pm0.014$
& $0.435\pm0.032$ & $0.110\pm0.021$
& $0.405\pm0.030$ & $0.177\pm0.035$
& $0.371\pm0.027$ & $0.313\pm0.045$
& $0.366\pm0.028$ & $0.388\pm0.053$ \\

& Flow Matching
& $0.046\pm0.028$ & $0.799\pm0.092$
& $0.359\pm0.020$ & $0.152\pm0.022$
& $0.356\pm0.022$ & $0.189\pm0.031$
& $0.346\pm0.021$ & $0.254\pm0.043$
& $0.332\pm0.022$ & $0.369\pm0.051$
& $0.324\pm0.021$ & $0.435\pm0.052$ \\

& \textbf{PIS}
& $\mathbf{0.016\pm0.009}$ & $\mathbf{0.859\pm0.142}$
& $\mathbf{0.022\pm0.013}$ & $\mathbf{0.699\pm0.060}$
& $\mathbf{0.029\pm0.016}$ & $\mathbf{0.537\pm0.089}$
& $\mathbf{0.034\pm0.016}$ & $\mathbf{0.436\pm0.097}$
& $\mathbf{0.041\pm0.019}$ & $\mathbf{0.337\pm0.110}$
& $\mathbf{0.042\pm0.019}$ & $\mathbf{0.336\pm0.111}$ \\
\midrule
\multirow{5}{*}{\makecell[l]{Wave-based\\Characterization\\(Helmholtz equations)}}
& PINN
& $0.157\pm0.039$ & $0.715\pm0.075$
& $0.456\pm0.067$ & $0.308\pm0.059$
& $0.502\pm0.063$ & $0.285\pm0.051$
& $0.512\pm0.064$ & $0.282\pm0.049$
& $0.515\pm0.064$ & $0.280\pm0.049$
& $0.517\pm0.065$ & $0.279\pm0.050$ \\

& FNO
& $0.023\pm0.010$ & $0.979\pm0.016$
& $0.433\pm0.066$ & $0.303\pm0.049$
& $0.408\pm0.063$ & $0.249\pm0.056$
& $0.423\pm0.060$ & $0.214\pm0.056$
& $0.463\pm0.054$ & $0.192\pm0.055$
& $0.466\pm0.049$ & $0.198\pm0.054$ \\

& NIO
& $0.090\pm0.079$ & $0.879\pm0.112$
& $0.470\pm0.037$ & $0.128\pm0.042$
& $0.500\pm0.040$ & $0.148\pm0.043$
& $0.523\pm0.039$ & $0.158\pm0.044$
& $0.521\pm0.036$ & $0.219\pm0.055$
& $0.522\pm0.035$ & $0.257\pm0.058$ \\

& Flow Matching
& $0.090\pm0.044$ & $0.816\pm0.098$
& $0.555\pm0.037$ & $0.099\pm0.034$
& $0.559\pm0.043$ & $0.114\pm0.040$
& $0.556\pm0.044$ & $0.157\pm0.041$
& $0.574\pm0.046$ & $0.201\pm0.054$
& $0.587\pm0.051$ & $0.222\pm0.054$ \\

& \textbf{PIS}
& $\mathbf{0.014\pm0.004}$ & $\mathbf{0.990\pm0.005}$
& $\mathbf{0.100\pm0.040}$ & $\mathbf{0.888\pm0.052}$
& $\mathbf{0.149\pm0.052}$ & $\mathbf{0.818\pm0.077}$
& $\mathbf{0.203\pm0.062}$ & $\mathbf{0.735\pm0.089}$
& $\mathbf{0.276\pm0.077}$ & $\mathbf{0.643\pm0.104}$
& $\mathbf{0.294\pm0.078}$ & $\mathbf{0.614\pm0.107}$ \\
\midrule
\multirow{5}{*}{\makecell[l]{Structural\\Health Monitoring}}
& PINN
& $0.115\pm0.010$ & $0.805\pm0.019$
& $0.854\pm0.085$ & $0.023\pm0.038$
& $0.793\pm0.071$ & $0.027\pm0.033$
& $0.776\pm0.072$ & $0.028\pm0.034$
& $0.766\pm0.068$ & $0.031\pm0.035$
& $0.762\pm0.067$ & $0.031\pm0.035$ \\

& FNO
& $0.020\pm0.002$ & $0.984\pm0.003$
& $0.895\pm0.097$ & $0.032\pm0.045$
& $0.854\pm0.090$ & $0.038\pm0.047$
& $0.847\pm0.089$ & $0.045\pm0.049$
& $0.842\pm0.089$ & $0.059\pm0.062$
& $0.841\pm0.089$ & $0.063\pm0.062$ \\

& NIO
& $0.045\pm0.043$ & $0.958\pm0.051$
& $0.990\pm0.087$ & $0.034\pm0.035$
& $1.050\pm0.084$ & $0.022\pm0.035$
& $1.034\pm0.099$ & $0.017\pm0.040$
& $0.985\pm0.102$ & $0.028\pm0.047$
& $0.975\pm0.099$ & $0.028\pm0.046$ \\

& Flow Matching
& $0.054\pm0.055$ & $0.947\pm0.065$
& $0.879\pm0.120$ & $0.075\pm0.061$
& $0.936\pm0.112$ & $0.053\pm0.053$
& $0.969\pm0.106$ & $0.035\pm0.051$
& $0.964\pm0.109$ & $0.044\pm0.058$
& $0.954\pm0.100$ & $0.046\pm0.062$ \\

& \textbf{PIS}
& $\mathbf{0.013\pm0.003}$ & $\mathbf{0.993\pm0.004}$
& $\mathbf{0.010\pm0.026}$ & $\mathbf{0.996\pm0.018}$
& $\mathbf{0.066\pm0.145}$ & $\mathbf{0.954\pm0.114}$
& $\mathbf{0.276\pm0.266}$ & $\mathbf{0.780\pm0.237}$
& $\mathbf{0.539\pm0.248}$ & $\mathbf{0.525\pm0.247}$
& $\mathbf{0.679\pm0.203}$ & $\mathbf{0.380\pm0.212}$ \\
\bottomrule
\end{tabular}}
\end{table*}

\textbf{Results.}
As detailed in Table~1, PIS achieves the lowest RMSE across all physical scenarios. In geometry-critical tasks (Helmholtz, SHM), it preserves phase information and local singularities, outperforming the strongest baselines by wide margins (e.g., SSIM $0.614$ vs. $0.222$). Notably, in subsurface inversion, NFM exhibits a ``hallucination'' effect---yielding competitive perceptual scores (SSIM) but high physical error (RMSE $0.324$)---whereas PIS maintains rigorous fidelity (RMSE $0.042$) and successfully reconstructs complex morphologies (e.g., conductivity channels, wave fronts) where baselines produce blurred artifacts, all while converging within a highly efficient 50 NFEs budget.

\subsection{Information and Sensor Efficiency Analysis}

\begin{figure}[ht]
  \centering 
  \includegraphics[width=0.8\columnwidth]{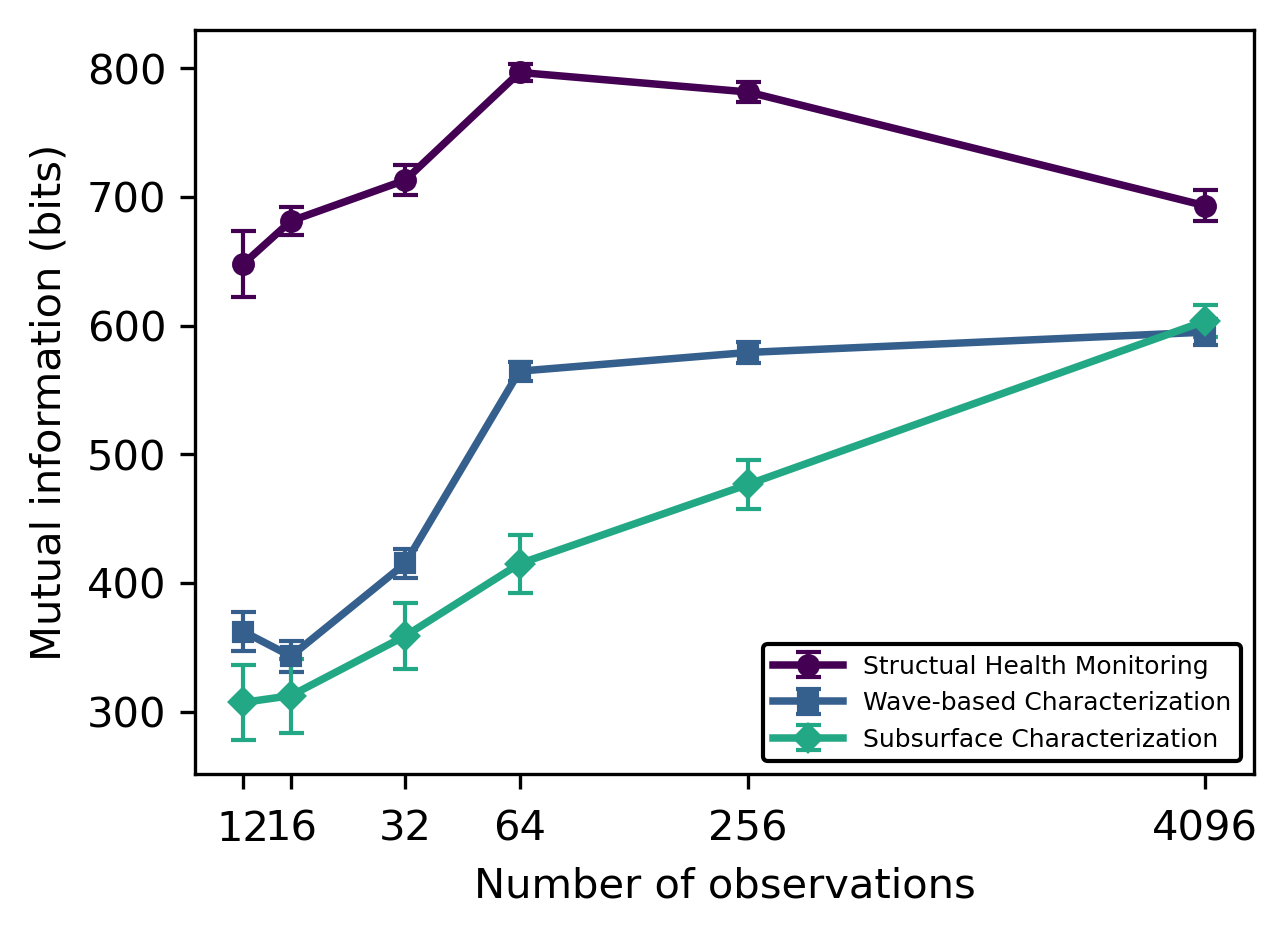}
  
  \caption{
  \textbf{Information Capacity.} Mutual information vs. observation budget. The curve shows rapid initial gain followed by saturation, highlighting the diminishing returns of adding sensors beyond a moderate budget.
  }
  \label{fig:IC}
\end{figure}

To evaluate the fundamental limits of sparse inversion, we quantify the Mutual Information (MI) between the latent field $\mathbf{x}$ and observations $\mathbf{Y}_{\omega}$ via entropy difference: $I(\mathbf{x}; \mathbf{y}_{\omega}) = H(\mathbf{x}) - H(\mathbf{x} \mid \mathbf{y}_{\omega})$ \cite{kaipio2005,chen2022integration}. We approximate these terms using the Kozachenko--Leonenko K-NN estimator, a robust non-parametric method for high-dimensional spaces. Specifically, estimates are derived from 200 unconditional samples (prior) and 200 conditional posterior samples generated by PIS at each sparsity budget.

\textbf{Results.}
The MI curves in Figure~\ref{fig:IC} reveal distinct scaling laws governed by physical constraints. Subsurface characterization exhibits near-linear growth, reflecting the multi-scale fractal nature of porous media where denser sensors continuously resolve finer heterogeneity without an early ceiling \cite{kitanidis1995}. Wave-based characterization displays a distinct ``phase transition'': information remains suppressed at low counts ($|\omega| \leq 16$) due to aliasing, but surges rapidly between 32 and 64, confirming a critical density is required to resolve oscillatory wavefronts \cite{woodward2007information}. SHM demonstrates the highest initial capacity ($>600$ bits) even with minimal observations but rapid saturation beyond $|\omega| = 64$, indicating that while global displacements are highly informative for damage detection, signal redundancy increases sharply with sensor density.

\textbf{PIS as a Tool for Optimal Sensor Placement.} Across all tasks, MI curves exhibit minor fluctuations due to randomness in sampled observation layouts. These variations reflect the sensitivity of each system to sensor placement and underscore a key strength of PIS: naturally enabling efficient simulation of alternative sensor layouts and quantitative evaluation based on information content. Furthermore, unlike traditional solvers that require re-training or expensive optimization for each new layout, PIS provides a zero-shot evaluation of information gain, positioning it as a practical tool for engineers to design configurations maximizing information gain under deployment constraints.

\subsection{Uncertainty Quantification}

Beyond deterministic estimation, PIS rigorously quantifies predictive risk by transporting a batch of noise samples ($K = 200$) to form a posterior ensemble. We conduct classical uncertainty and error analysis \cite{psaros2023uncertainty} with pixel-wise spatial analysis guided by calibrated regression framework \cite{kendall2017uncertainties}.

\begin{figure}[ht]
  \centering 
  \includegraphics[width=\columnwidth]{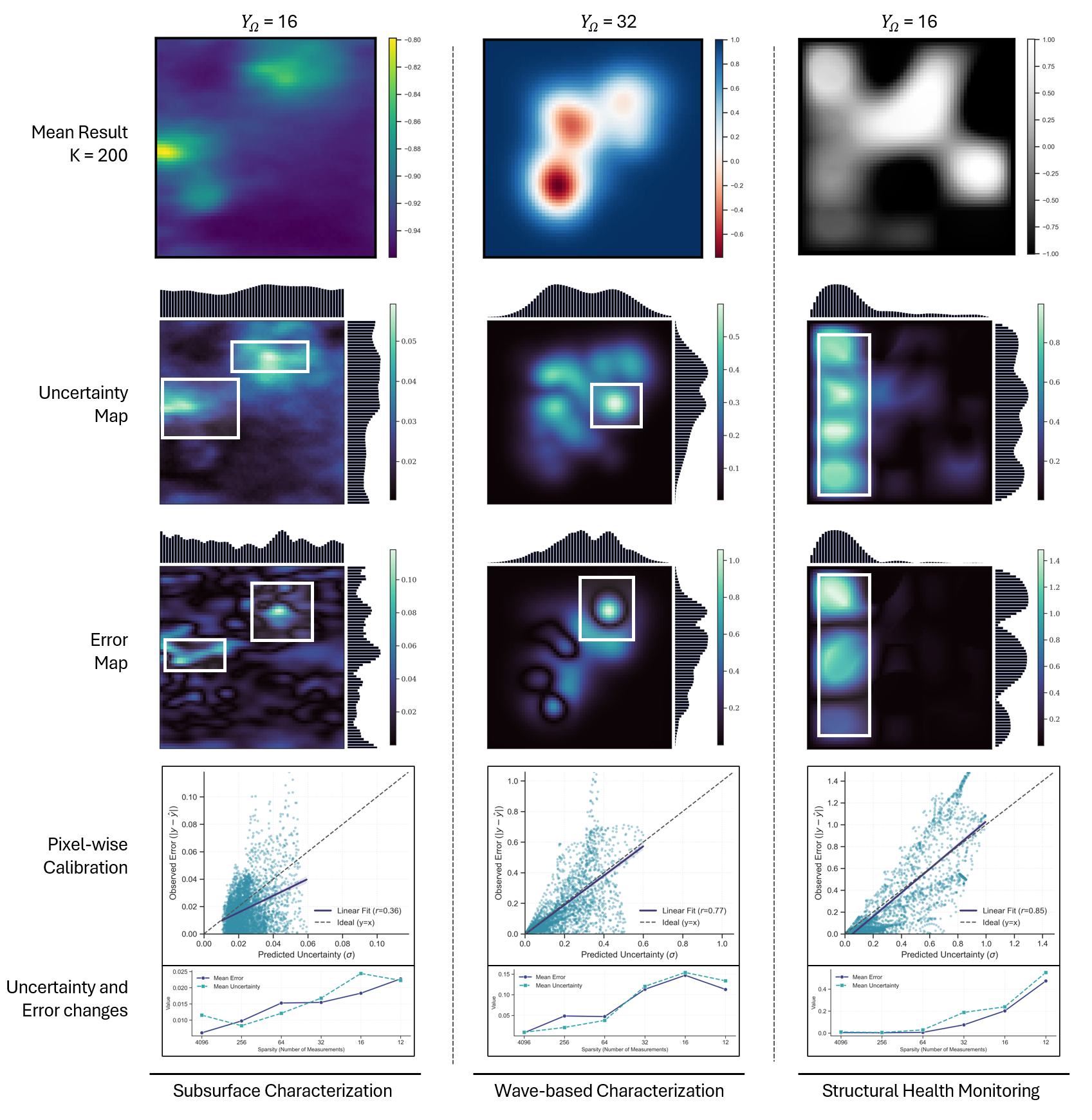}
  
  \caption{
  \textbf{Uncertainty Quantification.} Mean, variance, and error maps. Calibration plots confirm a strong uncertainty-error correlation, validating reliable estimation even under extreme sparsity.
  }
  \label{fig:UQ}
\end{figure}

\begin{figure*}[t]
    \centering
    \includegraphics[width=0.95\textwidth]{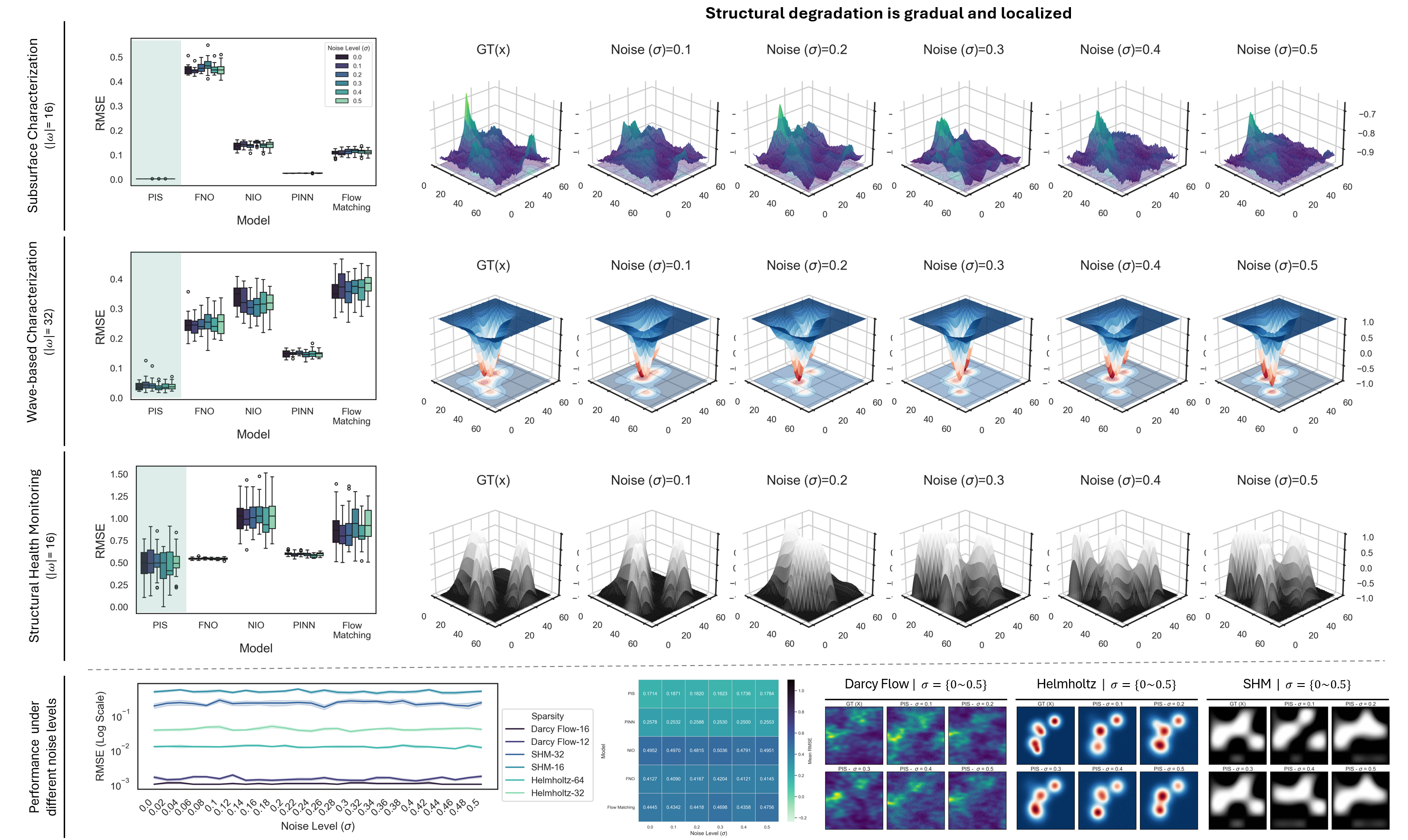}
    \caption{\textbf{Robustness to noise and arbitrariness across physical inverse problems.} Performance comparison under increasing Gaussian observation noise ($\sigma \in [0, 0.5]$) across three inverse problems. Left: RMSE distributions under fixed observation budgets guided by information analysis. Middle: Representative inversions at different noise levels. Right: Error maps highlighting noise-induced artifacts. PIS consistently degrades gracefully with noise and remains stable under arbitrary sensor locations.}
    \label{fig:noise}
\end{figure*}

\textbf{Results.} 
As shown in Figure~\ref{fig:UQ}, we observe:

\begin{itemize}
    \item \textbf{Spatial Alignment:} Uncertainty maps exhibit strong structural correspondence with error maps. High-variance regions consistently align with large residuals, demonstrating PIS's ``self-awareness'' to correctly identify failure-prone domains.
    
    \item \textbf{Calibration and Reliability:} Pixel-wise Calibration scatter plots reveal a robust positive correlation between predicted uncertainty and observed error, with Pearson coefficients $r$ ranging from $0.36$ to $0.85$. The strong correlation confirms that flow-derived variance is a reliable proxy for true predictive risk.
    
    \item \textbf{Sparsity Scaling:} Both global error and uncertainty scale monotonically with observation scarcity. Under extreme sparsity ($|\omega| = 12$), the posterior correctly diffuses, reflecting the necessary expansion of the solution space when physical constraints are minimal.
\end{itemize}

\subsection{Robustness to Noise and Arbitrariness}

In practical applications, observation data are frequently corrupted by noise due to sensor imprecision and environmental disturbances. Moreover, receivers are rarely distributed on regular grids. Therefore, robustness to both measurement noise and arbitrary observation patterns is essential \cite{colton1998inverse}. Following established benchmarks \cite{cao2025entropy}, we inject Gaussian noise with standard deviation $\sigma_{\text{noise}} \in [0, 0.5]$ into the random observation samples. We evaluate the RMSE variations across 20 independent generations for each noise level to quantify the stability of PIS against its baselines.

\textbf{Results.} Across all three scenarios (Figure~\ref{fig:noise}, left column), PIS maintains the most stable and minimum RMSE distributions even as noise increases. While baselines exhibit volatile RMSE fluctuations and significant performance degradation under high noise, PIS's error remains tightly constrained. This highlights the effectiveness of the SCTU-Net's conditional mechanism, which leverages the learned physical prior to ``regularize'' the noisy likelihood signal during the flow integration process. In addition, PIS also demonstrates visual fidelity under corruption. Even under strong corruption ($\sigma_{\text{noise}} = 0.5$), its inversions preserve global topological structures and fine-grained details. For example, in subsurface characterization, the hydraulic conductivity channel structure remains connected and sharp.

\textbf{PIS as a Foundation for Sim-to-Real Transfer.} The strong anti-noise and layout-agnostic capabilities position PIS as a powerful tool for transfer learning. Since PIS can intake an arbitrary number of receivers with variable placements and high noise levels, it bridges the ``sim-to-real'' gap. Researchers can pre-train PIS on large-scale synthetic datasets and fine-tune it with scarce real-world measurements.

\subsection{Ablation Studies}

Table~\ref{tab:ablation} isolates the contribution of core components. CASC and the Set Transformer prove paramount; their removal triggers catastrophic degradation in the extreme sparsity regime ($|\omega|=12$), confirming their essential role in stabilizing vector fields from unstructured inputs. Global and Spatial conditioning offer complementary benefits: the former dictates the physical regime at moderate sparsity, while the latter anchors local geometry as observations dwindle. These results validate the synergy between PIS's adaptive curriculum and its multi-scale set architecture.
\begin{table}[h]
\caption{\textbf{Ablation Study.} We report the RMSE averaged across three physical regimes under varying sparsity levels. Lower is better.}
  \label{tab:ablation}
  \centering 
  \scriptsize
  \begin{sc}
    \setlength{\tabcolsep}{9pt}
    \begin{tabular}{lcccc}
      \toprule
      Variant & 64 & 32 & 16 & 12 \\
      \midrule
      Full PIS & \textbf{0.081} & \textbf{0.171} & \textbf{0.285} & \textbf{0.338} \\
      w/o Set Trans. & 0.168 & 0.302 & 0.464 & 0.521 \\
      w/o Global Cond. & 0.105 & 0.221 & 0.358 & 0.429 \\
      w/o Spatial Cond. & 0.098 & 0.208 & 0.324 & 0.391 \\
      w/o Curriculum & 0.143 & 0.361 & 0.612 & 0.787 \\
      \bottomrule
    \end{tabular}
  \end{sc}
\end{table}

\section*{Conclusion}
\label{sec:conclusion}
We introduced the Physical Inversion Solver (PIS), a novel generative framework for high-fidelity physical inversion under extreme and unstructured observation layouts. By integrating the SCTU-Net with Flow Matching, PIS overcomes the rigid grid dependencies and high computational overhead in existing methods. Extensive experiments across subsurface characterization, wave-based characterization, and structural health monitoring demonstrate that PIS maintains remarkable structural integrity even when observations cover less than $0.3\%$. The CASC ensures stable convergence in highly ill-posed regimes, effectively bridging the informational gap.

Beyond inversion, PIS serves as a powerful tool for optimal sensor placement. Its zero-shot evaluation capability for arbitrary sensor layouts allows engineers to quantify information gain instantaneously. By reducing sampling curvature through straight-path probability flows, PIS achieves orders-of-magnitude speedup, providing a scalable foundation for real-time diagnostics and sim-to-real transfer. We provide a detailed discussion of limitations, including resolution scalability and noise sensitivity, in Appendix~G. Future work will extend PIS to time-dependent PDEs and high-resolutions, further broadening its applicability to dynamic systems.

\clearpage
\section*{Impact Statement}

This paper presents Physical Inversion Solver (PIS), a generative framework designed to solve high-dimensional inverse problems in physics and engineering. The primary societal impact of this work lies in its potential to enhance the efficiency and reliability of monitoring systems in critical domains, including groundwater resource management, geophysical wave exploration,structural health monitoring, as well as subsurface applications such as oil and gas extraction and geological carbon storage. By enabling high-fidelity inversion from sparse and arbitrary sensor layouts, our method can contribute to more sustainable environmental practices and safer infrastructure maintenance, reducing the economic and logistical costs of data acquisition. Compared to multi-stage inversion pipelines that require substantial theoretical expertise and careful system design, PIS consolidates the inference procedure into a single unified model, streamlining implementation and integration with existing monitoring workflows in engineering and industrial environments.

However, we acknowledge that deploying data-driven generative models in safety-critical applications carries inherent risks, particularly the possibility of model hallucination or overconfidence in inversion, which could lead to misinformed decisions (e.g., failing to detect structural damage). To mitigate these risks, our work explicitly integrates and evaluates Uncertainty Quantification (UQ) mechanisms. We strongly advocate that practitioners utilize these UQ metrics to assess predictive confidence before making high-stakes decisions. We do not foresee this work facilitating harmful surveillance, discrimination, or other negative ethical consequences common in general physical inversion tasks.

\nocite{langley00}
\clearpage
\bibliography{example_paper}
\bibliographystyle{icml2026}

\newpage
\appendix
\onecolumn
\section{Related Work}
\label{sec:related}

\subsection{Classical and Bayesian Physical Inversion}
Classical inverse problem theory dates to the foundational work of Hadamard \cite{hadamard1902,hadamard1923}, with early geophysical formulations outlined in \cite{parker1977,tarantola1982}. In classical approaches, physical inversion is typically carried out using deterministic optimization schemes such as Tikhonov-regularized least squares, Gauss--Newton and Levenberg--Marquardt iterations, and adjoint-state methods for gradient computation. These approaches require repeated evaluations of the forward operator $\mathcal{G}$ and its adjoint, which are themselves implemented through computationally intensive forward PDE solvers such as finite-difference (FDM), finite-element (FEM), and spectral methods. Because the underlying problems are often severely ill-posed, additional regularization---including total variation (TV) \cite{rudin1992tv}, Laplacian smoothing, or sparsity-promoting penalties---is required to stabilize the inversion. Such regularization inevitably introduces model-dependent bias into the solution \cite{engl1996regularization,kaipio2005,tarantola1987inverse}.

Bayesian inversion frameworks provide a principled alternative \cite{tarantola1987inverse,kaipio2005}, with rigorous infinite-dimensional formulations developed in \cite{stuart2010inverse,buithanh2014pde}. Posterior inference in these models typically relies on Markov Chain Monte Carlo (MCMC), Hamiltonian Monte Carlo, variational inference, or ensemble-based approaches, all of which require thousands of forward or adjoint PDE solves \cite{neal2012011mcmc}. While such techniques yield full posterior uncertainty, their computational cost limits applicability to large-scale, high-dimensional fields. In many engineering and geoscience applications, observational data are inherently sparse, expensive to acquire, and constrained to irregular or nonstandard sensor locations due to physical accessibility, economic and environmental constraints. Recent surveys in geophysical inversion \cite{zhao2024bvi} emphasize that sparse, irregular, or geometry-dependent acquisition further exacerbates instability and increases ambiguity, requiring stronger regularization or more samples. However, all of these classical approaches fundamentally depend on repeated evaluations of the forward or adjoint PDE operator and deteriorate rapidly when observations become sparse, irregular, or highly undetermined. This leaves open the need for scalable inversion frameworks that can operate reliably under severe observational sparsity without requiring repeated PDE solves.

\subsection{Information-Theoretic Perspectives}
\label{subsec:info_theory}

Information theory provides principled tools for quantifying the informativeness of measurements. Shannon entropy \cite{shannon1948} and related quantities such as mutual information \cite{cover2006,lindley1956} have been widely used to study identifiability, experimental design, and data value in inverse problems \cite{alexanderian2014}. These ideas motivate our analysis of how observational sparsity limits posterior contraction and how different physical systems exhibit distinct information--sparsity curves.

In the specific domain of hydraulic conductivity, these information-theoretic metrics have emerged as a cornerstone for optimal Monitoring Network Design (MND) and uncertainty reduction. Unlike traditional variance-based criteria, entropy-based methods effectively capture the high-order statistics of non-Gaussian subsurface heterogeneity \cite{mogheir2002application}. Recent advances have further synergized information theory with deep learning surrogates to tackle high-dimensional inverse problems. For instance, Chen et al.\ (2022) \cite{chen2022integration} proposed an integrated framework combining deep learning with the Maximum Information Minimum Redundancy (MIMR) criterion to optimize sensor placement, significantly reducing uncertainty in heterogeneous aquifers. Building on this, Cao et al.\ (2025a) \cite{cao2025integrated} utilized entropy theory to guide high-dimensional permeability field identification, demonstrating that entropy-guided sampling outperforms random strategies in capturing complex geological features. More recently, Cao et al.\ (2025b) \cite{cao2025entropy} extended this paradigm to multivariate network design, leveraging entropy to balance the information gain from multi-source observations (e.g., hydraulic heads and solute concentrations) for robust data assimilation.

These ideas motivate our analysis within the PIS framework: we leverage these foundational entropy concepts not just to design static networks, but to quantitatively analyze how observational sparsity fundamentally limits posterior contraction across distinct physical systems.

\subsection{Deep Learning and Diffusion Models for Physical Inversion}
\label{subsec:deep_learning}

\textbf{Discriminative Models.} Deep learning has been extensively applied to physical inversion, ranging from convolutional encoder-decoder architectures for seismic waveform inversion \cite{jin2017deep,wu2019inversionnet} to physics-informed neural networks (PINNs) for solving inverse PDEs \cite{raissi2019pinn}. Recent advances in operator learning, such as DeepONet \cite{lu2021deeponet}, Fourier Neural Operators (FNO) \cite{li2021fourier}, and Neural Inverse Operators (NIO) \cite{molinaro2023neural}, have demonstrated significant potential in mapping measurements to physical fields. However, these models typically assume fixed-grid representations and provide single-point deterministic predictions. They often lack the robustness required for highly sparse or irregularly distributed measurements and fail to quantify posterior uncertainty---a critical requirement in severely underdetermined regimes.

\textbf{Diffusion and Score-based Models.} These limitations have motivated a shift toward generative modeling: denoising diffusion probabilistic models (DDPM) \cite{ho2020ddpm} and score-based models \cite{song2020score} have shown strong ability to capture complex distributions, and have been adopted for inverse problems via Diffusion Posterior Sampling (DPS) \cite{chung2023dps} or related likelihood-guided samplers \cite{song2021score,song2023pseudoinverse}, as well as end-to-end conditional diffusion frameworks for subsurface inversion such as geological CO\textsubscript{2} storage property characterization \cite{wang2025generative}. While effective, these approaches typically assume a known, differentiable forward operator and rely on dense, structured conditioning. Furthermore, the high sampling curvature of diffusion-based SDEs necessitates hundreds of iterations, leading to prohibitive computational costs for real-time physical monitoring.

\textbf{The Rise of Flow Matching.} Recently, Flow Matching (FM) \cite{lipman2022flow,liu2023flow} has emerged as a superior alternative to diffusion by learning to regress deterministic velocity fields. By inducing a straight-line probability flow between noise and data, FM significantly reduces sampling curvature and enables efficient inference with minimal function evaluations. Despite its success in image synthesis, the application of FM to physical inversion under arbitrary, off-grid observational constraints remain underexplored.

\subsection{Set-Based Learning for Unordered and Arbitrary Sparse Observations}
\label{subsec:set_based}

A complementary line of work studies learning from unordered, variable-sized collections of inputs, meeting the model design requirement of an arbitrary number of observation inputs. DeepSets \cite{zaheer2017deepsets} introduced permutation-invariant neural architectures, while Set Transformer \cite{lee2019settransformer} further leverages self-attention to capture interactions within sets. Neural Processes and Attentive Neural Processes \cite{garnelo2018neural,kim2019attentive} extend these ideas to probabilistic function regression, modelling distributions conditioned on context sets. Despite their flexibility, these models are not designed for PDE-governed physical inversion. They typically operate in low-dimensional function spaces, lack mechanisms to enforce physical constraints, and do not integrate generative processes, such as diffusion, to produce posterior samples. Moreover, existing set-based methods generally assume that the target quantities lie in simple function spaces, whereas physical inversion requires reconstructing high-dimensional fields governed by complex physics. These set-based architectures provide the flexibility to encode arbitrary, unordered, and variable-sized observation sets, which is essential for handling sparse and irregular measurements. Despite these limitations, their architectural flexibility remains attractive for developing scalable observation encoders in generative physical inversion frameworks.

\subsection{Uncertainty Quantification (UQ) in Physical Inversion}
\label{subsec:uq}

Quantifying uncertainty is essential in ill-posed inverse problems, where multiple solutions may fit the available data. Classical approaches to uncertainty quantification rely primarily on Bayesian sampling techniques such as Markov chain Monte Carlo and Hamiltonian Monte Carlo \cite{neal2012011mcmc} or Gaussian process--based models \cite{rasmussen2006gp}. While theoretically well-founded, these methods become computationally prohibitive in high-dimensional PDE-governed settings due to the need for repeated forward or adjoint solves. To improve scalability, approximate Bayesian deep learning methods have been proposed, including MC dropout \cite{gal2016dropout}, deep ensembles \cite{lakshminarayanan2017deepensembles}, and Bayesian neural networks \cite{blundell2015weightuncertainty}. Although more tractable, these approaches typically provide limited calibration, capture only coarse epistemic uncertainty, and remain tied to deterministic forward mappings that do not model the full posterior distribution. Recent neural operator--based UQ methods \cite{molinaro2023neural} extend these ideas to PDE settings, but they still produce deterministic predictions and do not incorporate generative sampling. Importantly, most existing UQ methods assume dense or structured observations and are not designed to handle highly sparse or irregular measurement configurations. As a result, developing uncertainty-aware inversion frameworks that remain reliable under severe observational sparsity remains an open challenge.

\clearpage
\section{Implementation Details}
\label{sec:implementation}
\subsection{PIS Pseudo-Algorithm}
\label{subsec:algorithm}
\begin{algorithm}[ht]
  \caption{PIS Training with Cosine-Annealed Sparsity Curriculum (CASC)}
  \label{alg:pis_fm_casc}
  \begin{algorithmic}
    \STATE \textbf{Input:} Dataset of physical fields $\mathcal{D}=\{x^{(i)}\}$, sparsity range $[|\omega|_{\min}, |\omega|_{\max}]$, training iterations $K$
    \STATE \textbf{Output:} Optimized SCTU-Net parameters $\theta$
    \STATE Initialize SCTU-Net $v_\theta$ and Field Synthesis (FS) module parameters
    \FOR{$k = 1$ \textbf{to} $K$}
        \STATE Sample physical field $x_1 \sim \mathcal{D}$
        \STATE \textbf{Step 1: Curriculum Sparsity Selection}
        \STATE Compute sparsity budget using cosine annealing:
        \[
            |\omega|_k = |\omega|_{\min} 
            + \frac{1}{2}\left(|\omega|_{\max} - |\omega|_{\min}\right)
            \left(1 + \cos\left(\frac{k\pi}{K}\right)\right)
        \]
        \STATE Randomly sample $|\omega|_k$ indices $\omega \subset \Omega$
        \STATE Extract sparse observations: $y_\omega = x_1[\omega] + \varepsilon$
        \STATE \textbf{Step 2: Flow Matching}
        \STATE Sample time $t \sim \text{Uniform}(0,1)$ and noise $x_0 \sim \mathcal{N}(0, I)$
        \STATE Construct intermediate state: $x_t = t x_1 + (1 - t)x_0$
        \STATE Define target velocity: $u_t(x_t) = x_1 - x_0$
        \STATE \textbf{Step 3: Optimization}
        \STATE Extract set-conditioned features: $z_\omega = \text{FS}(\text{SCT}(y_\omega))$
        \STATE Predict velocity field: $\hat{v}_t = v_\theta(x_t, t, z_\omega)$
        \STATE Update $\theta$ by minimizing:
        \[
            \mathcal{L}_{\text{total}}
            = \mathbb{E}_{t, x_0, x_1}
            \left\| \hat{v}_t - u_t \right\|^2
            + \lambda \, \mathrm{Var}(\hat{v}_t)
        \]
    \ENDFOR
    \STATE \textbf{Return} optimized parameters $\theta$
  \end{algorithmic}
\end{algorithm}

\begin{algorithm}[H]
  \caption{PIS Inference (Posterior Sampling)}
  \label{alg:pis_fm_inference}
  \begin{algorithmic}
    \STATE \textbf{Input:} Sparse observations $y_\omega$ at arbitrary locations $\omega$, trained model $v_\theta$, number of steps $N$
    \STATE \textbf{Output:} Reconstructed field $\hat{x}_1$ and uncertainty map $\sigma$
    \STATE \textbf{Pre-processing:}
    \STATE Encode observations into latent features: $z_\omega = \text{FS}(\text{SCT}(y_\omega))$
    \STATE Initialize noise samples $\{x_0^{(i)}\}_{i=1}^{M} \sim \mathcal{N}(0, I)$
    \FOR{$i = 1$ \textbf{to} $M$}
        \STATE \textbf{ODE Integration:}
        \STATE Define dynamics:
        \[
            \frac{d x_t}{d t} = v_\theta(x_t, t, z_\omega),
            \quad x(0) = x_0^{(i)}
        \]
        \STATE Solve for $x_1^{(i)}$ using an ODE solver (Euler with $N$ steps)
    \ENDFOR
    \STATE \textbf{Post-processing:}
    \STATE Compute mean inversion:
    \[
        \hat{x}_1 = \frac{1}{M} \sum_{i=1}^{M} x_1^{(i)}
    \]
    \STATE Compute uncertainty map:
    \[
        \sigma = \mathrm{Std}\left(\{x_1^{(i)}\}_{i=1}^{M}\right)
    \]
    \STATE \textbf{Return} $\hat{x}_1$, $\sigma$
  \end{algorithmic}
\end{algorithm}

\subsection{Hyperparameters and Implementation Details}
\label{subsec:hyperparams}
All models are implemented in PyTorch and trained on a single NVIDIA A100 (80GB) GPU. The implementation is fully differentiable, allowing for end-to-end training of the set-conditioned flow matching objective.

\subsubsection{Architecture Configuration}
\label{subsec:arch_config}

The SCTU-Net serves as the backbone for the velocity field parameterization $v_{\theta}$. It consists of two specialized encoders and a conditional U-Net decoder.

\textbf{Observation Encoders (SCT).} We employ two parallel Set Transformers to extract complementary features from the input set $\mathbf{y}_{\omega} \in \mathbb{R}^{|\omega| \times D_{\text{in}}}$.

\begin{itemize}
    \item \textbf{Architecture:} Each encoder follows the sequence: \textsc{IASB} $\rightarrow$ \textsc{IASB} $\rightarrow$ \textsc{PMA} $\rightarrow$ \textsc{SAB} $\rightarrow$ \textsc{SAB}. We utilize $M = 32$ inducing points in ISAB layers to maintain linear complexity $\mathcal{O}(M \cdot |\omega|)$.
    
    \item \textbf{Configuration:} The encoders operate with a hidden dimension of $D_{\text{hidden}} = 256$ and $H = 8$ attention heads.
    
    \item \textbf{Dual-Path Outputs:} The Spatial Encoder utilizes $k = 64$ seed vectors in the PMA layer to produce location-aware features aligned with the spatial grid, while the Global Encoder uses $k = 1$ seed vector to extract a global context embedding of dimension $D_{\text{global}} = 512$.
\end{itemize}

\textbf{Field Decoder (U-Net).} The decoder is a modified U-Net with $N_{\text{res}} = 3$ residual blocks per resolution level.

\begin{itemize}
    \item \textbf{Dimensions:} It employs channel multipliers of $[1, 2, 4, 8]$ with a base channel width of 64. The bottleneck features interact with the global context via 16-head cross-attention mechanisms.
    
    \item \textbf{Conditioning:} Time embeddings $t$ are injected into every residual block via MLP projections (SiLU activated).
\end{itemize}

\subsubsection{Training Protocol}
\label{subsec:training_protocol}

\textbf{Optimization.} We train PIS using the Adam optimizer with standard betas $\beta_1 = 0.9$, $\beta_2 = 0.999$ and no weight decay. Gradient clipping is set to $1.0$ to ensure stability.

\textbf{Learning Rate Schedule.} The learning rate is initialized at $\eta = 1 \times 10^{-4}$ and follows a Cosine Annealing Warm Restarts schedule.

\textbf{Two-Stage Training.}
\begin{itemize}
    \item \textbf{Stage 1 (Warmup):} The model is trained on dense observations ($|\omega| = 4096$) for 300 epochs to learn the global physics prior.
    
    \item \textbf{Stage 2 (Curriculum):} The model is fine-tuned for 1000 epochs using the CASC strategy, where sparsity varies dynamically.
\end{itemize}

\textbf{Batch Size.} A batch size of 32 is used across all tasks.

\subsubsection{Flow Matching and Sparsity Curriculum}
\label{subsec:flow_matching_details}

\begin{table}[t]
  \caption{Hyperparameter configuration for PIS implementation.}
  \label{tab:hyperparams}
  \begin{center}
    \begin{small}
      \begin{sc}
        \begin{tabular}{llc}
          \toprule
          Component & Hyperparameter & Value \\
          \midrule
          \multirow{5}{*}{Optimization}
          & Optimizer & Adam \\
          & Learning Rate ($\eta$) & $1 \times 10^{-4}$ \\
          & Batch Size & 32 \\
          & Training Epochs & 300 (Stage 1) + 1000 (Stage 2) \\
          & Gradient Clip & 1.0 \\
          \midrule
          \multirow{4}{*}{SCT Encoder}
          & Hidden Dimension & 256 \\
          & Attention Heads & 8 \\
          & Inducing Points ($M$) & 32 \\
          & Layer Structure & $2 \times \text{IASB} + 1 \times \text{PMA} + 2 \times \text{SAB}$ \\
          \midrule
          \multirow{6}{*}{U-Net Backbone}
          & Global Feature Dim & 512 \\
          & Base Channels & 64 \\
          & Channel Multipliers & $[1, 2, 4, 8]$ \\
          & ResBlocks per Level & 3 \\
          & Attention Heads & 16 \\
          & Dropout & 0.1 \\
          \midrule
          \multirow{4}{*}{Flow Matching}
          & Training $t$ Distribution & $\mathcal{U}(0,1)$ \\
          & Noise Distribution & $\mathcal{N}(0,1)$ \\
          & Variance Reg ($\lambda$) & 0.001 \\
          & Inference Solver & Euler (50 steps) \\
          \bottomrule
        \end{tabular}
      \end{sc}
    \end{small}
  \end{center}
\end{table}

\textbf{Objective Function.} We minimize the conditional flow matching loss with an optional variance regularization term ($\lambda_{\text{var}} = 0.001$) enabled during the sparsity curriculum phase to prevent mode collapse. The complete loss is given by:

\begin{equation}
\mathcal{L}_{\text{total}} = \mathbb{E}_{t, \mathbf{x}_0, \mathbf{x}_1, \mathbf{y}_{\omega}} \left[ \| v_{\theta}(\mathbf{x}_t, t, \mathbf{y}_{\omega}) - (\mathbf{x}_1 - \mathbf{x}_0) \|^2 \right] + \lambda_{\text{var}} \cdot \text{Var}(v_{\theta})
\end{equation}

where $\text{Var}(v_{\theta})$ computes the variance of the predicted velocity field across the batch dimension.

\textbf{CASC Schedule Details.} 
The sparsity curriculum dynamically anneals the observation count from a dense regime ($|\omega|_{\max} = 4096$) down to the target sparsity ($|\omega|_{\min} = 12$) following a cosine trajectory over the total training iterations $K$. The schedule is defined as:

\begin{equation}
|\omega|_k = |\omega|_{\min} + \frac{1}{2} \left( |\omega|_{\max} - |\omega|_{\min} \right) \left( 1 + \cos\left(\frac{k\pi}{K}\right) \right)
\end{equation}

where $k$ denotes the current training step and $K$ is the total number of optimization steps. 
This schedule ensures that the model initially learns from high-information, well-posed states (dense observations) and gradually adapts to the ill-posed, sparse regime, effectively performing continuation on the optimization landscape.

\textbf{Inference Configuration.} For sampling, we solve the probability flow ODE using the Euler method with a standard budget of NFE $= 50$ (unless otherwise specified for efficiency ablation). The integration follows:

\begin{equation}
\mathbf{x}_{t+\Delta t} = \mathbf{x}_t + \Delta t \cdot v_{\theta}(\mathbf{x}_t, t, \mathbf{y}_{\omega}), \quad \Delta t = \frac{1}{\text{NFE}}
\end{equation}

starting from $\mathbf{x}_0 \sim \mathcal{N}(\mathbf{0}, \mathbf{I})$ and integrating to $t = 1$ to obtain the posterior sample $\mathbf{x}_1$.

\subsubsection{Observation Set Construction Strategy}
\label{subsec:obs_set_construction}

To rigorously evaluate the model's capability to handle unstructured data, we construct the observation sets $\mathbf{y}_{\omega}$ through a stochastic sampling process that mimics real-world sensor deployment.

\paragraph{Stochastic Sampling Protocol.}
For a given ground-truth physical field $\mathbf{u} \in \mathbb{R}^{H \times W}$, we define a continuous domain $\Omega = [0,1]^2$. The observation set is generated as follows:

\begin{itemize}
    \item \textbf{Cardinality Sampling:} For each training instance, we first sample the number of sensors $K \sim \mathcal{U}(|\omega|_{\min}, |\omega|_{\max})$ (e.g., from 12 to 4096).
    
    \item \textbf{Coordinate Selection:} We sample $K$ unique spatial coordinates $\mathbf{s}_i = (x_i, y_i)$, which are then normalized to the continuous range $[0, 1]$.
    
    \item \textbf{Value Extraction:} The observations are extracted as $v_i = \mathbf{u}(x_i, y_i)$.
    
    \item \textbf{Set Formatting:} The final input is formatted as an unordered set of vectors:
    \begin{equation}
        \mathbf{y}_{\omega} = \{(\mathbf{s}_i, v_i)\}_{i=1}^{K} \in \mathbb{R}^{K \times D_{\text{in}}}
    \end{equation}
\end{itemize}

\paragraph{Normalization and Standardization.}
To ensure stable training dynamics for the Flow Matching objective, we apply strict normalization:

\begin{itemize}
    \item \textbf{Coordinates:} Spatial coordinates are normalized to $[0, 1]$.
    
    \item \textbf{Observation Values:} The values $v_i$ are standardized to an approximate range of $[-1, 1]$ using domain-specific statistics (e.g., log-permeability for Darcy flow and advection-diffusion equations, displacement magnitude for SHM). This matching of scales between coordinates and values is crucial for the SCTU-Net to learn balanced attention weights.
\end{itemize}

\subsubsection{Baseline Architectures}
\label{app:baselines}

To ensure a fair and rigorous comparison, all baseline models are implemented using PyTorch and tuned to match the parameter count of our proposed PIS where applicable. Detailed configurations are as follows:

\textbf{1. Fourier Neural Operator (FNO).}
We adopt the 2D FNO architecture. The model consists of 4 Fourier layers with a lifting dimension of 128. In each layer, we retain the top 12 frequency modes in both spatial directions. The projection layer utilizes GeLU activation. The total parameter count is approximately 2.4M.

\textbf{2. Amortized Physics-Informed Neural Networks (Amortized PINN).}
Unlike standard PINNs that require per-instance optimization, we employ an amortized framework to enable fast inference. The model learns a generalized mapping using a U-Net architecture. The training objective minimizes $\mathcal{L} = \mathcal{L}_{\text{data}} + \lambda_{\text{PDE}}\mathcal{L}_{\text{physics}}$, allowing the model to predict solutions for unseen sensor inputs via a single forward pass.

\textbf{3. Neural Inverse Operator (NIO).}
Designed to directly approximate the inverse mapping, the NIO baseline employs a DeepONet-based architecture \cite{molinaro2023neural}. A Branch Net (MLP) encodes the sparse sensor observations, while MLP encodes the query coordinates. The final field is reconstructed via the dot product of the two feature vectors. Both networks consist of 4 hidden layers with 128 units, trained via supervised MSE loss.

\textbf{4. Standard Flow Matching (NFM).}
This baseline represents a vanilla application of Optimal Transport Flow Matching without set-specific architectural biases. 
\begin{itemize}
    \item \textbf{Backbone:} It utilizes a standard U-Net (identical to the decoder of PIS but without cross-attention to set features).
    \item \textbf{Input Processing:} Since standard U-Nets cannot process unstructured sets, sparse off-grid observations are pre-processed via linear interpolation to construct a dense $64 \times 64$ grid input.
    \item \textbf{Training and Inference:} It is trained with the same objective and sampled using an Euler solver with NFE = 50.
\end{itemize}
Comparing NFM with PIS isolates the contribution of our SCTU-Net architecture (which natively handles sets) versus simply feeding interpolated grids to a generative flow model.

\section{Physical Scenario and Datasets}
\label{sec:datasets}

\subsection{Subsurface Characterization}
\label{subsec:subsurface}

\paragraph{Physical Governing Equations.}
This section briefly presents the governing equations describing the dynamics of solute contaminant transport by groundwater flow within an aquifer. The governing equation for steady-state flow is expressed as follows:

\begin{equation}
\frac{\partial}{\partial x_i} \left( K_i \frac{\partial h}{\partial x_i} \right) = 0 \quad \text{in } \Omega
\end{equation}

After the flow equation is solved, the flow velocity $v_i$ can be calculated based on Darcy's Law with knowledge of the hydraulic head field:

\begin{equation}
v_i = -\frac{K_i}{\theta} \frac{\partial h}{\partial x_i}
\end{equation}

where $K_i$ denotes the hydraulic conductivity; $h$ is the hydraulic head; $x_i$ represents the Cartesian coordinate along two principal directions; and $\theta$ signifies the aquifer porosity. The transport of a conservative solute within the two-dimensional groundwater flow field is then described by the following advection--dispersion equation:

\begin{equation}
\frac{\partial (\theta C)}{\partial t} = \frac{\partial}{\partial x_i} \left( \theta D_{ij} \frac{\partial C}{\partial x_j} \right) - \frac{\partial}{\partial x_i} (\theta v_i C) + q_a c_s \quad \text{in } \Omega
\end{equation}

where $C$ is the solute concentration of contaminant; $t$ denotes time; $q_a$ is the flow rate of the source/sink term and $c_s$ is the concentration of source/sink term; $D_{ij}$ is the hydrodynamic dispersion tensor determined by the pore space flow velocity $v_i$. Under prescribed initial and boundary conditions and given a specified spatial distribution of hydraulic conductivity ($K$), the model responses can be computed by solving Equations~(15)--(17). Numerical groundwater flow and transport simulators such as MODFLOW \cite{MODFLOW} and MT3DMS \cite{MT3DMS} are commonly employed to solve these governing equations.

\paragraph{Inversion Target.}
The objective is to estimate the heterogeneous hydraulic conductivity field $\ln K$ based on sparse, spatiotemporal measurements of hydraulic head and solute concentration $\mathbf{y}_{\omega} = \{(h(\mathbf{x}_i), C(\mathbf{x}_i, t_j))\}_{(\mathbf{x}_i, t_j) \in \Omega}$.

\paragraph{Data Generation.}
\textit{Site Conceptualization and Setup:} This study examines a two-dimensional, steady-state groundwater flow scenario within a confined aquifer. The hypothetical aquifer covers a spatial area of $192 \times 192$ m and is discretized into a grid comprising 64 rows and 64 columns, resulting in uniform grid cells of $1.5 \times 1.5$ m. Impermeable boundary conditions are imposed along the northern and southern edges of the domain, while constant hydraulic heads of 11 m and 10 m are maintained along the western and eastern boundaries, respectively. The initial hydraulic head is uniformly set to 10 m throughout the domain, except at the western boundary where the specified head condition is applied.

\paragraph{K Field Generation.} The $\ln K$ field is randomly generated using the Karhunen--Loève Expansion (KLE) \cite{zhang2004efficient}. The spatial correlation structure of the log-conductivity is characterized by the following exponential covariance function:

\begin{equation}
\gamma(\mathbf{l}, \mathbf{l}')_{\ln K} = \sigma_{\ln K}^2 \exp\left(-\sqrt{\left(\frac{l_x - l_x'}{\lambda_x}\right)^2 + \left(\frac{l_y - l_y'}{\lambda_y}\right)^2}\right)
\end{equation}

where $\mathbf{l} = (l_x, l_y)$ and $\mathbf{l}' = (l_x', l_y')$ denote two arbitrary spatial locations, $\sigma_{\ln K}^2$ is the variance, $\lambda_x$ and $\lambda_y$ are the correlation lengths along $x$ and $y$ directions, respectively. Length scales of $\lambda_x/L_x = \lambda_y/L_y = 0.2$ are considered in this study, where $L_x$ and $L_y$ are size of flow domain along the $x$ and $y$ directions, respectively. The mean of $\ln K$ field $\mu_{\ln K}$ is 2.0, while the variances are set to $\sigma_{\ln K}^2 = 0.5$. The parameterization formula for the KLE can be expressed as follows:

\begin{equation}
\ln K(x, y) \approx \overline{\ln K}(x, y) + \sum_{i=1}^{N_{\text{KL}}} \xi_i \sqrt{\tau_i} s_i(x, y)
\end{equation}

where $\overline{\ln K}(x, y)$ is the mean of the $\ln K(x, y)$, $\xi_i$ are independent standard Gaussian random variables, $\tau_i$ and $s_i(x, y)$ are eigenvalues and eigenfunctions of the correlation function, respectively, and $N_{\text{KL}}$ is the number of KLE terms. In this study, the first 400 KLE terms were retained, capturing 97\% of the heterogeneity characteristics of the log-conductivity field:

\begin{equation}
\frac{\sum_{i=1}^{400} \tau_i}{\sum_{i=1}^{\infty} \tau_i} \approx 97.0\%
\end{equation}

\paragraph{Dataset.} We employ the KLE to generate a diverse ensemble of $\ln K$, which serve as the heterogeneous parameters to be estimated. Numerical simulations are performed using industry-standard solvers MODFLOW (for flow) and MT3DMS (for transport) with a time step $\Delta t = 50$ days, and the simulation terminates at $t = 1000$ days. Concentration data are recorded at time steps $t = [50:50:1000]$, along with hydraulic head measurements. For each spatial grid point (pixel), the observation vector consists of a single steady-state hydraulic head measurement and a time-series of solute concentration recorded at 20 distinct time steps. This process yields a dataset of 5,200 paired samples $(\ln K, \mathbf{y}_{\omega})$, which are randomly partitioned into 5,000 samples for training and 200 for testing.

\subsection{Wave-based Characterization}
\label{subsec:helmholtz}

\paragraph{Physical Governing Equations.} The Helmholtz equation is a fundamental equation in wave-based imaging, describing the propagation of waves, such as sound or electromagnetic waves, in a medium. In its general form, the Helmholtz equation for a scalar wavefield $u(\mathbf{x})$ is given by:

\begin{equation}
\nabla^2 u(\mathbf{x}) + k^2(\mathbf{x}) u(\mathbf{x}) = 0
\end{equation}

where $u(\mathbf{x})$ is the complex-valued wavefield, and $k(\mathbf{x})$ is the spatially varying wave number that encodes the local material properties of the medium, such as the refractive index or density.

\paragraph{Physical Setup.} In wave inversion problems, the wave number $k(\mathbf{x})$ is a key parameter that varies across the medium. It is related to the material properties of the medium, such as density $\rho(\mathbf{x})$ and the refractive index $n(\mathbf{x})$, with the following general form:

\begin{equation}
k(\mathbf{x}) = \frac{2\pi}{\lambda(\mathbf{x})}
\end{equation}

where $\lambda(\mathbf{x})$ is the spatially varying wavelength. For a medium with varying properties, this equation allows us to model the spatially dependent wave inversion accurately. The goal in inverse problems is to recover the wave number $k(\mathbf{x})$ based on sparse measurements of the wavefield $u(\mathbf{x})$, typically measured at certain sensor locations \cite{colton1998inverse}.

\paragraph{Solution of the Helmholtz Equation.} To solve the Helmholtz equation for wave inversion, we typically apply numerical methods such as the finite element method (FEM) or finite difference methods (FDM). These methods discretize the spatial domain and solve the partial differential equation (PDE) at each point. In this case, we use the public Helmholtz dataset from Hugging Face (\texttt{camlab-ethz/Helmholtz}), with each sample of a full resolution $64 \times 64$.

The boundary conditions required for the solution depend on the physical setup. For example, in a wave inversion scenario, the boundary conditions may correspond to the type of waves being modeled, such as Dirichlet or Neumann conditions, which represent fixed wave amplitudes or fixed wave gradients at the boundaries, respectively.

Once the wavefield $u(\mathbf{x})$ is obtained for a known $k(\mathbf{x})$, partial measurements of the wavefield are used to infer the unknown spatially varying wave number $k(\mathbf{x})$. This inversion process is a key challenge in wave-based imaging, as it is inherently ill-posed---meaning that there are multiple possible solutions to the inverse problem, and additional regularization or constraints are needed to obtain meaningful results \cite{aki2002quantitative}.

\paragraph{Inversion Target.} The inverse problem formulation involves estimating the unknown wave number field $k(\mathbf{x})$ from partial, sparse observations of the wavefield $u(\mathbf{x})$. This inverse problem can be framed as a minimization problem:

\begin{equation}
\hat{k}(\mathbf{x}) = \arg\min_k \| \mathcal{G}(k) - \mathbf{y} \|_2^2
\end{equation}

where $\mathcal{G}(k)$ is the forward operator that models the relationship between the wave number $k(\mathbf{x})$ and the observed wavefield, $\mathbf{y}$ represents the sparse observations (e.g., pointwise measurements of the wavefield), and $\|\cdot\|_2^2$ denotes the squared $L_2$-norm, measuring the difference between the observed data and the predicted wavefield. Solving this problem requires techniques for handling the ill-posed nature of the inverse problem, typically by incorporating regularization strategies or probabilistic models that account for uncertainty in the observations.

\subsection{Structural Health Monitoring}
\label{subsec:shm}

We investigate a static inverse elasticity problem representative of structural health monitoring (SHM) and Non-Destructive Evaluation (NDE). In engineering practice, identifying the spatial distribution of material stiffness (Young's modulus $E(\mathbf{x})$) is critical for detecting internal defects, assessing material degradation, or characterizing heterogeneous composites. However, deploying dense sensor arrays on large-scale structures is often economically or physically infeasible. Consequently, the challenge lies in reconstructing the high-dimensional stiffness field from sparse displacement measurements obtained at limited sensor locations.

\paragraph{Physical Governing Equations.} We model the structure as a two-dimensional continuum governed by linear elasticity under plane strain conditions. The mechanical equilibrium is defined by the conservation of linear momentum:

\begin{equation}
\nabla \cdot \bm{\sigma} = \mathbf{0} \quad \text{in } \Omega
\end{equation}

subject to Dirichlet boundary conditions. The stress tensor $\bm{\sigma}$ relates to the linearized strain tensor $\bm{\epsilon} = \frac{1}{2}(\nabla \mathbf{u} + (\nabla \mathbf{u})^T)$ through the isotropic constitutive law:

\begin{equation}
\sigma_{ij} = \frac{E}{1+\nu} \left( \epsilon_{ij} + \frac{\nu}{1-2\nu} \epsilon_{kk} \delta_{ij} \right)
\end{equation}

where $\mathbf{u}$ denotes the displacement vector field, and $\nu$ is the Poisson's ratio. The inverse problem aims to infer the spatially varying scalar field $E(\mathbf{x})$ solely from a sparse observation set $\mathbf{y}_{\omega} = \{\mathbf{u}(\mathbf{x}_i)\}_{\mathbf{x}_i \in \Omega}$.

\paragraph{Data Generation.} To simulate a rigorous engineering NDE scenario, we employ a Finite Element Method (FEM) pipeline to generate physically consistent datasets.

\begin{itemize}
    \item \textbf{FEM Simulation:} The forward mechanical response is solved using scikit-fem. The domain is defined as a unit square representative volume element (RVE). Consistent with standard load testing procedures in SHM, we fix the bottom boundary and apply a uniform 1\% compressive strain ($u_y = -0.01$) to the top boundary. We set the Poisson's ratio $\nu = 0.40$ \cite{bastek2023inverse}, simulating the behavior of nearly incompressible materials such as elastomeric isolators or rubber-toughened composites.
    
    \item \textbf{Stiffness Field Construction $E(\mathbf{x})$:} Following the Function Sampling Space (FSS) \cite{ni2021deep} framework, we model the domain as a two-phase heterogeneous medium consisting of a matrix and randomly distributed stiff inclusions. To generate topologically diverse and realistic material gradients, we adopt a cubic spline-based generation strategy. We first initialize a $4 \times 4$ latent grid where values are randomly assigned to either a matrix phase ($E_{\text{matrix}} = 10.0$) or an inclusion phase ($E_{\text{inclusion}} = 50.0$), representing a high-contrast ratio of 5:1 typically found in hard inclusions or tumors. This coarse grid is then upsampled to a $64 \times 64$ resolution using cubic spline interpolation to create smooth, realistic transitions between materials.
\end{itemize}

\section{Evaluation Metrics and ODE Solvers}
\label{sec:evaluation}

\subsection{RMSE (Root Mean Square Error)}
\label{subsec:rmse}

RMSE (Root Mean Square Error) quantifies global inversion error and acts as the main metric \cite{chai2014rmse}:

\begin{equation}
\text{RMSE} = \sqrt{\frac{1}{N} \sum_{i=1}^{N} \| \hat{\mathbf{x}}^{(i)} - \mathbf{x}^{(i)} \|_2^2}
\end{equation}

\subsection{SSIM (Structural Similarity Index)}
\label{subsec:ssim}

SSIM (Structural Similarity Index) measures perceptual and spatial consistency for inversion field \cite{wang2004ssim}:

\begin{equation}
\text{SSIM}(u,v) = \frac{(2\mu_u \mu_v + C_1)(2\sigma_{uv} + C_2)}{(\mu_u^2 + \mu_v^2 + C_1)(\sigma_u^2 + \sigma_v^2 + C_2)}
\end{equation}

where $\mu_u, \mu_v$ are the local means, $\sigma_u^2, \sigma_v^2$ are the variances, $\sigma_{uv}$ is the covariance, and $C_1, C_2$ are small constants for numerical stability.

\subsection{ODE Solver}
\label{subsec:ode_solver}

During inference, generating physical field inversions from the learned posterior involves numerically integrating the neural Probability Flow ODE defined in Section~3. Starting from a sampled noise vector $\mathbf{x}_0 \sim \mathcal{N}(\mathbf{0}, \mathbf{I})$ at time $t = 0$, we solve for the data state $\mathbf{x}_1$ at $t = 1$ following the trajectory dictated by the learned velocity field $v_{\theta}$:

\begin{equation}
d\mathbf{x}_t = v_{\theta}(\mathbf{x}_t, t, \mathbf{z}_{\omega}) dt, \quad t \in [0,1]
\end{equation}

where $\mathbf{z}_{\omega} = \text{FS}(\text{SCT}(\mathbf{y}_{\omega}))$ is the set-conditioned latent representation held constant during integration.

\paragraph{Euler Integration Scheme.} While higher-order solvers (e.g., Runge--Kutta--45) are often necessary for standard diffusion models due to their highly curved reverse-SDE trajectories, our use of the Optimal Transport Flow Matching objective induces straight integration paths between the noise and data distributions. This geometric property significantly reduces discretization errors, allowing us to employ the computationally efficient first-order Euler method. We discretize the time interval $[0,1]$ into $N$ uniform steps, yielding a time grid $\{t_0, t_1, \dots, t_N\}$ where $t_i = i/N$ and the step size is $\Delta t = 1/N$. The iterative update rule is given by:

\begin{equation}
\mathbf{x}_{t_{i+1}} = \mathbf{x}_{t_i} + v_{\theta}(\mathbf{x}_{t_i}, t_i, \mathbf{z}_{\omega}) \Delta t, \quad \text{for } i = 0, \dots, N-1
\end{equation}

\paragraph{Implementation.} For all main experiments reported in Table~1, we use $N = 50$ steps (NFE $= 50$). This choice strikes an optimal balance between inversion fidelity and computational efficiency, offering an order-of-magnitude speedup compared to standard iterative solvers that typically require hundreds of steps. We observed negligible performance gains when increasing $N$ beyond 50, confirming the low-curvature nature of the learned flow.

\clearpage
\section{Experiments Supplements and Additional Results}
\label{sec:supplment}

\subsection{Inversion Quality for wave-based characterization and SHM}
\label{sec:supplment_e1}

\begin{figure}[H]
  \centering
  
  \includegraphics[width=0.85\columnwidth]{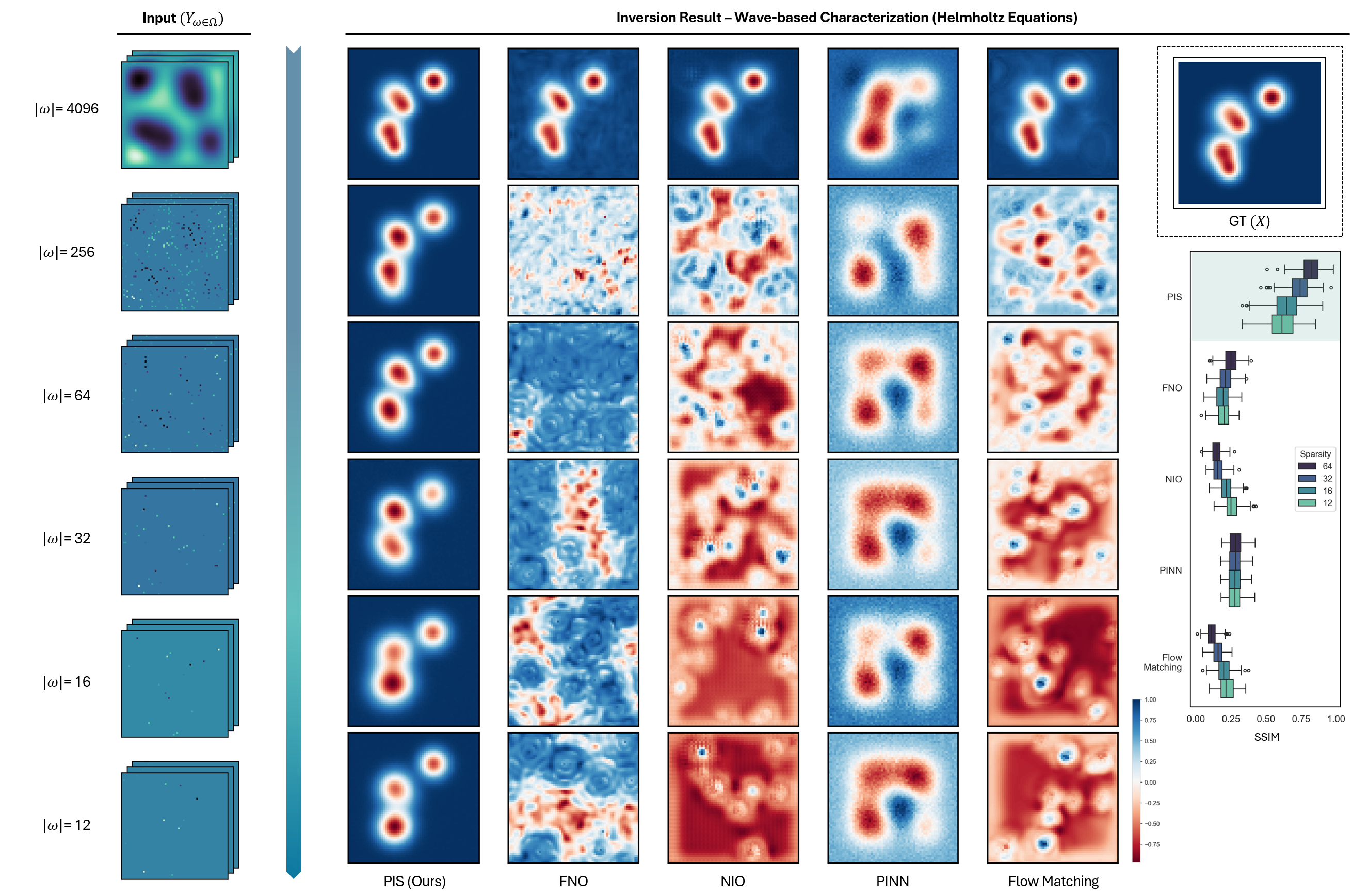}
  \caption{}
  \label{fig:top}

  \includegraphics[width=0.85\columnwidth]{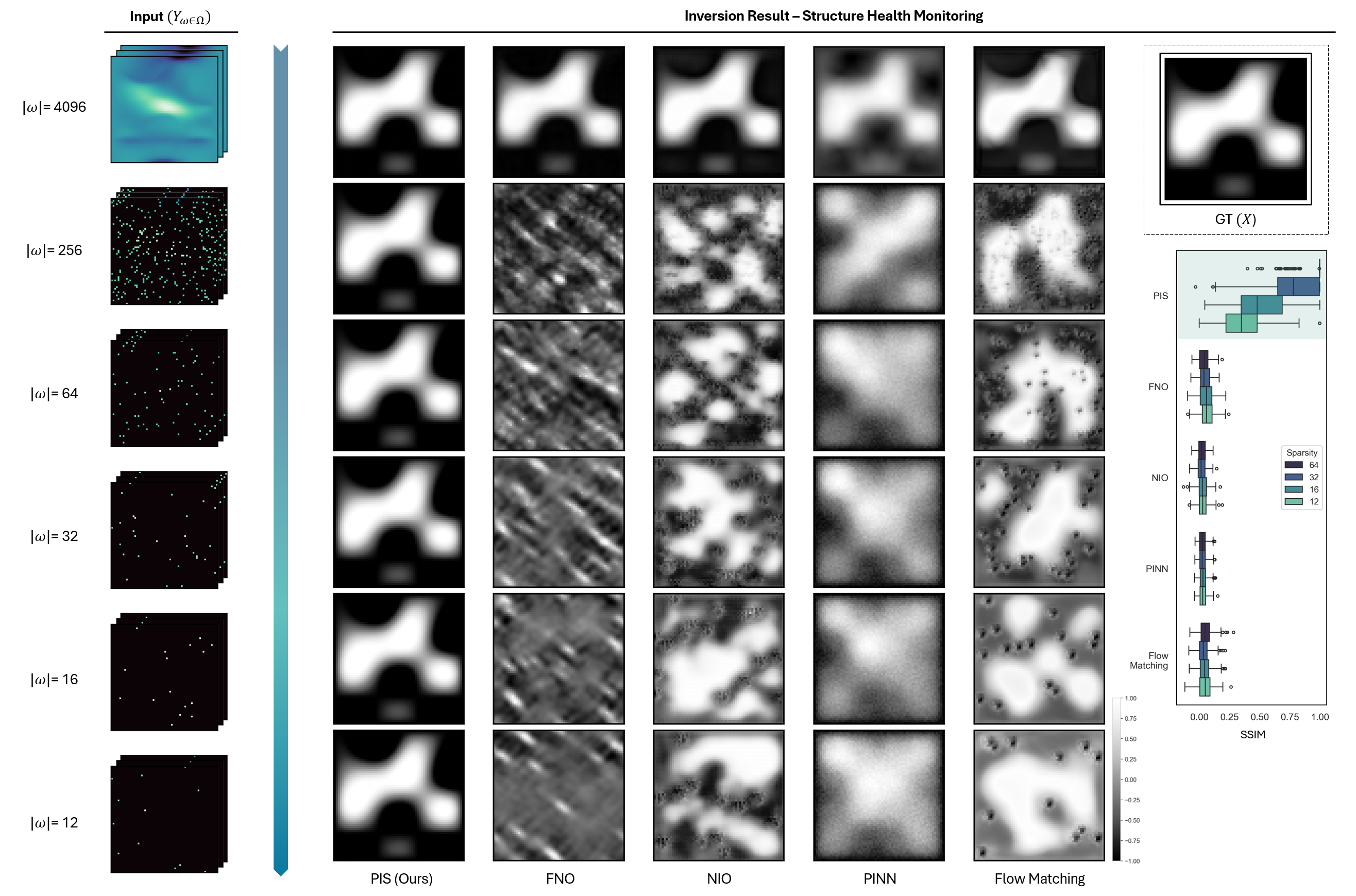}
  \caption{}
  \label{fig:bottom}

\end{figure}

\clearpage
\subsection{Uncertainty Quantifications under multiple observation sparsity}
\label{sec:supplment_e2}
\begin{figure}[H]
  \centering
  
  \includegraphics[width=0.95\columnwidth]{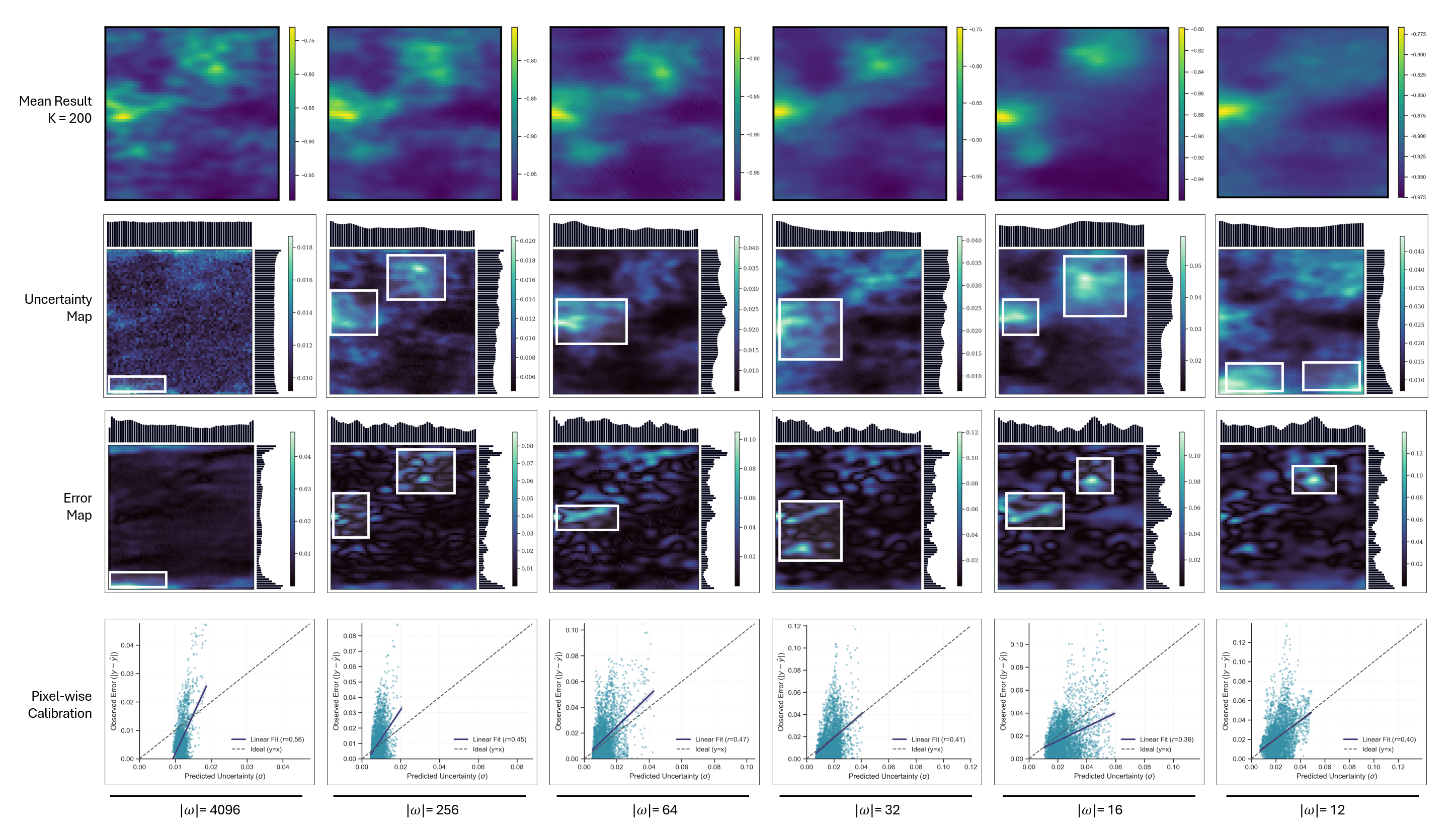}
  \caption{}
  \label{fig:top}

  \includegraphics[width=0.95\columnwidth]{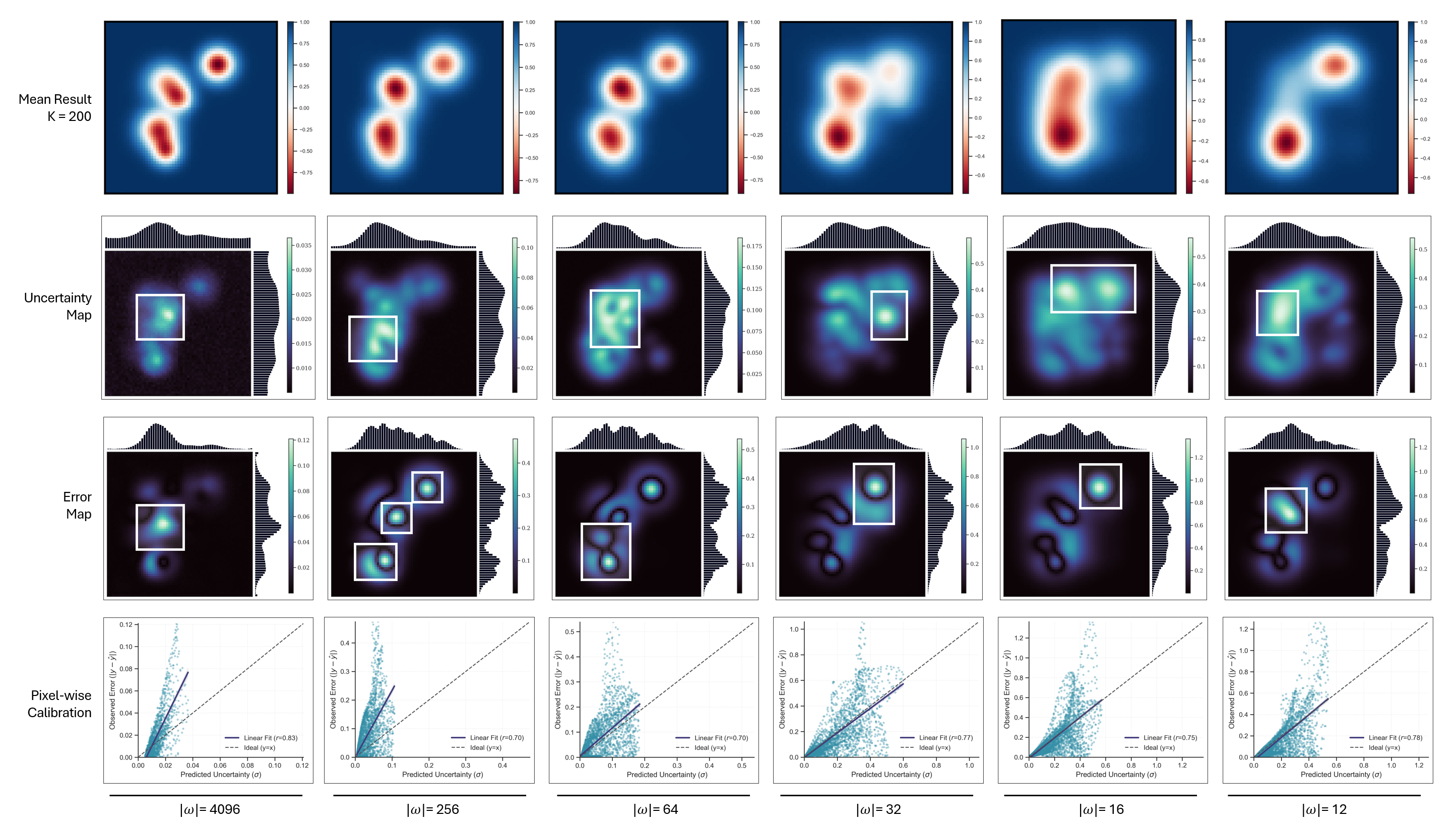}
  \caption{}
  \label{fig:middle}

\end{figure}

\clearpage
\begin{figure}[H]
  \centering
  \includegraphics[width=0.95\columnwidth]{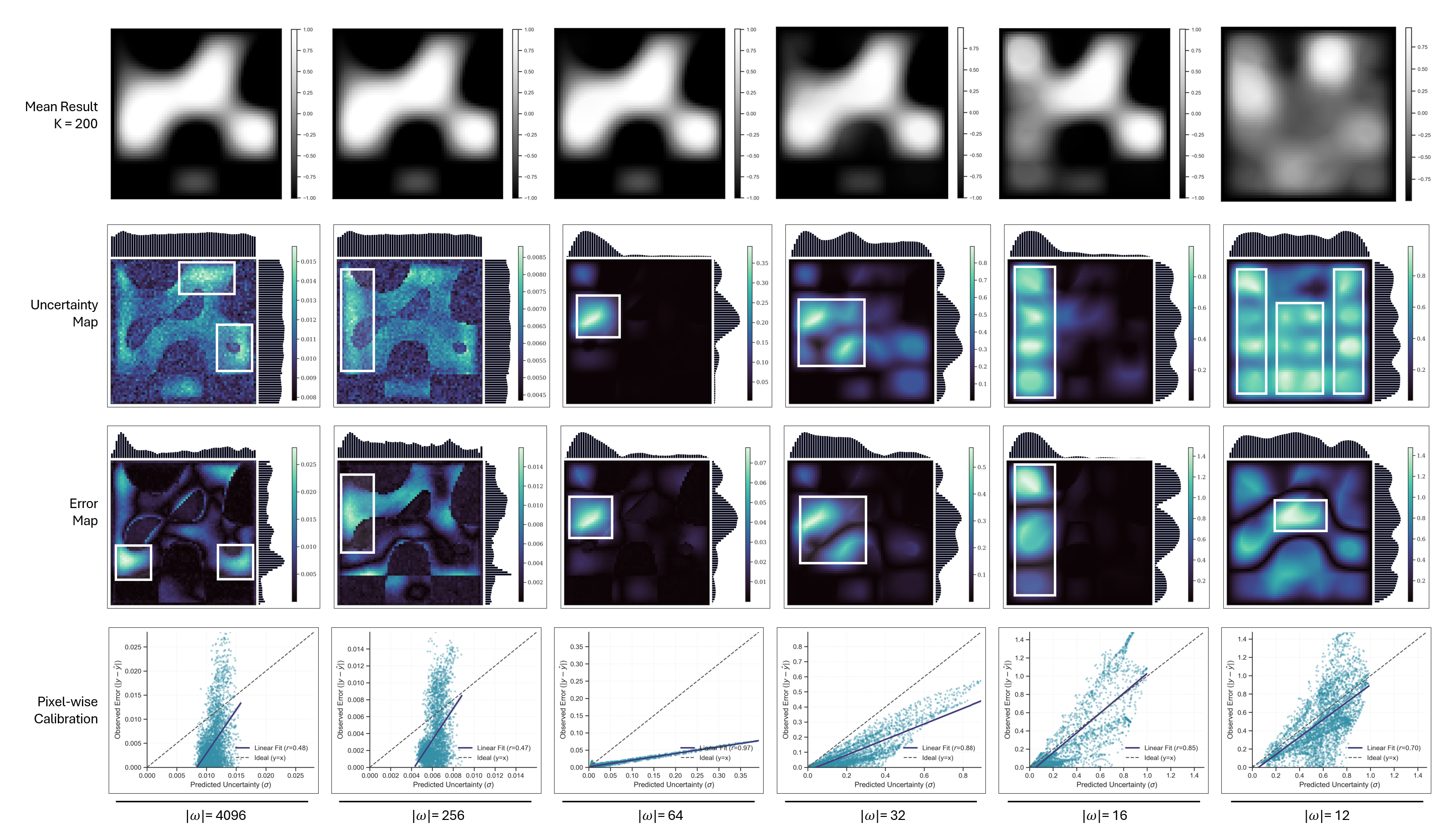}
  \caption{}
  \label{fig:top}
\end{figure}

\clearpage
\section{PIS Framework Comparison: Flow Matching and DDPM}
\label{sec:comparison}
\subsection{Inversion Quality Analysis}
\label{subsec:fm_vs_ddpm}
\begin{figure}[H]
  \centering
  \includegraphics[width=0.95\columnwidth]{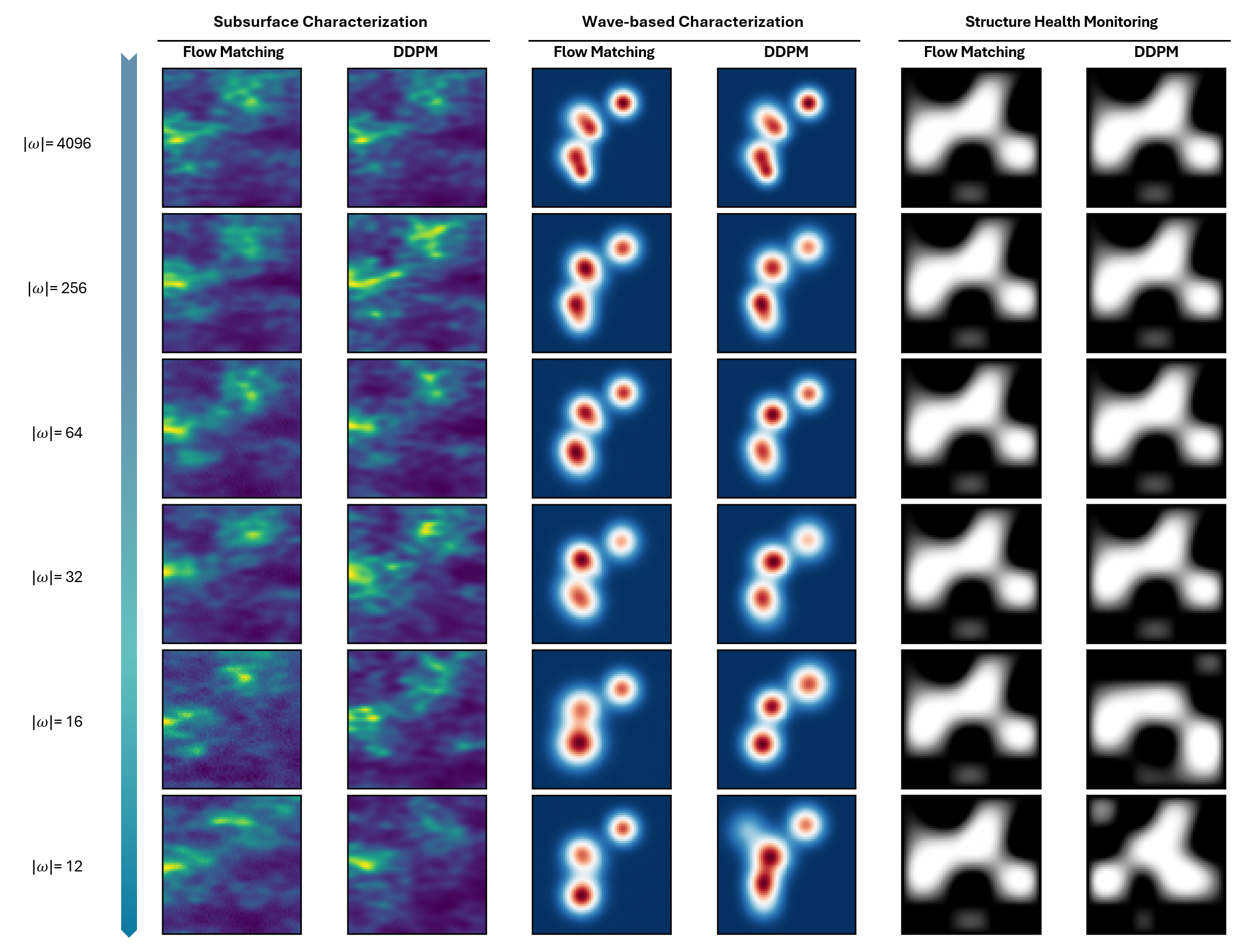}
  \caption{}
  \label{fig:top}
\end{figure}
To demonstrate the architectural flexibility and robustness of the PIS framework, we implemented and evaluated PIS using two distinct generative paradigms: Conditional Flow Matching (FM) and Denoising Diffusion Probabilistic Models (DDPM). Both variants share the same SCTU-Net backbone and CASC training strategy.

\paragraph{Visual Fidelity and Structural Consistency.}
As illustrated in Figure~11, both PIS and PIS-DDPM demonstrate remarkable capability in capturing the posterior distribution of the physical fields across all three tasks.

\begin{itemize}
    \item \textbf{Subsurface Characterization:} Both models resolve high-permeability channels accurately, even at $|\omega| = 12$.
    \item \textbf{Wave-based Characterization:} Both paradigms avoid phase artifacts and maintain wave-front sharp boundaries.
    \item \textbf{SHM:} The localized stiffness inclusions are consistently identified by both models.
\end{itemize}

The high degree of visual similarity between the two frameworks confirms that the Set-Conditioned architecture and Curriculum Learning strategy are the primary drivers of PIS's robustness to extreme sparsity, independent of the underlying generative sampling mechanism.

\paragraph{Efficiency and Straight-Path Advantage.}
While both models achieve comparable inversion quality, PIS exhibits a significant practical advantage in inference efficiency.

\begin{itemize}
    \item \textbf{Sampling Curvature:} DDPM relies on stochastic reverse SDEs which often require a high number of sampling steps to maintain stability in highly under-determined regimes.
    \item \textbf{Speedup:} In contrast, Flow Matching leverages the straight-path property of the learned vector field, enabling high-fidelity inversion within significantly fewer function evaluations (50 NFEs).
\end{itemize}

\paragraph{Conclusion.} These results validate that the PIS framework is operator-agnostic and generalizes effectively across different generative families. While DDPM remains a reliable alternative, Flow Matching is our preferred implementation for real-time engineering applications due to its superior efficiency and deterministic probability flow.

\subsection{Computation Complexity Analysis}
\label{subsec:complexity}

\subsubsection{Backbone Complexity}
Both Flow Matching and DDPM for PIS share the same SCTU-Net backbone. Let $H,W$ denote the spatial dimensions, $K$ the number of sparse observations ($K = |\omega|$), and $M,k_{\text{seed}}$ the number of inducing points and seed vectors, respectively. The complexity is analyzed as follows:

\begin{itemize}
    \item \textbf{Set-Conditioned Transformer (SCT):} The code implementation employs a specialized ISAB-PMA-SAB architecture that decouples the computational cost from the observation count $K$.
    \begin{itemize}
        \item \textit{Linear Aggregation (ISAB \& PMA):} The encoder uses Induced Set Attention Blocks (ISAB) and Pooling Multihead Attention (PMA) to process the input set. These operations scale linearly as $\mathcal{O}(K \cdot (M + k_{\text{seed}}) \cdot D)$, where $M = 32$ and $k_{\text{seed}} = 64$ are small constants. This ensures efficient handling of dense observations ($K = 4096$).
        \item \textit{Constant Refinement (SAB):} Crucially, the subsequent Self-Attention Blocks (SAB) operate solely on the aggregated seed vectors ($k_{\text{seed}}$). Their complexity is $\mathcal{O}(k_{\text{seed}}^2 \cdot D)$, which is constant with respect to input sparsity $K$.
        \item \textit{Conclusion:} The total SCT complexity is $\mathcal{O}(K \cdot M \cdot D + k_{\text{seed}}^2 \cdot D)$, making it strictly linear w.r.t. $K$ and computationally negligible compared to grid-based operations.
    \end{itemize}
    
    \item \textbf{U-Net Backbone:} The convolution-based U-Net layers scale linearly with the spatial resolution, $\mathcal{O}(H \cdot W \cdot C_{\text{feat}})$.
    
    \item \textbf{Total Backbone Complexity:} $C_{\text{backbone}} \approx \mathcal{O}(M \cdot K \cdot D + H \cdot W \cdot C_{\text{feat}})$.
\end{itemize}

\subsubsection{Training Complexity}
In the training phase, both paradigms require one forward and one backward pass per iteration.

\begin{itemize}
    \item \textbf{DDPM:} Optimizes the variational lower bound or a reweighted score-matching loss, necessitating the estimation of the score function $\nabla \log p_t(\mathbf{x})$.
    \item \textbf{Flow Matching:} Directly regresses the constant-velocity vector field $u_t(\mathbf{x}) = \mathbf{x}_1 - \mathbf{x}_0$.
\end{itemize}

\paragraph{Conclusion.} The per-iteration training complexity is nearly identical for both: $\mathcal{O}(C_{\text{backbone}})$. However, Flow Matching often exhibits more stable convergence due to its deterministic, straight-path probability flow, which simplifies the optimization landscape.

\subsubsection{Inference Complexity}
\label{subsubsec:inference_complexity}

This is the primary differentiator between the two methods. The total inference complexity $C_{\text{inf}}$ is a function of the Number of Function Evaluations (NFE): $C_{\text{inf}} = \text{NFE} \times C_{\text{backbone}}$.

\begin{itemize}
    \item \textbf{PIS-DDPM (SDE-based):} Due to the high curvature of the stochastic reverse-time SDE, standard diffusion solvers typically require $\text{NFE}_{\text{DDPM}} \in [100, 1000]$ steps to ensure sampling stability and avoid cumulative discretization errors.
    
    \item \textbf{PIS (ODE-based):} By constructing a straight-path vector field where $\mathbf{x}_t$ is a linear interpolation between noise and data, PIS induces a flow with minimal curvature. This allows the use of large step sizes in deterministic ODE solvers (e.g., Euler or RK45), achieving convergence in $\text{NFE}_{\text{FM}} \in [40, 50]$ steps.
\end{itemize}

\subsubsection{Theoretical Speedup}
\label{subsubsec:speedup}

The theoretical speedup factor $S$ of Flow Matching over DDPM under PIS is given by:

\begin{equation}
S = \frac{\text{NFE}_{\text{DDPM}} \cdot C_{\text{backbone}}}{\text{NFE}_{\text{FM}} \cdot C_{\text{backbone}}} = \frac{\text{NFE}_{\text{DDPM}}}{\text{NFE}_{\text{FM}}}
\end{equation}

Given our experimental configurations, $S$ ranges from $10\times$ to $50\times$. This efficiency gain is achieved without compromising the expressive power of the generative posterior, positioning PIS as a more scalable solution for real-time physical inversion.

\section{Limitations}
\label{sec:limitations}

While PIS demonstrates significant advantages in efficiency, robustness, and handling arbitrary sparsity, we acknowledge several limitations that provide avenues for future research.

\subsection{Dimensionality and Geometric Complexity}
\label{appendix:limitations_resolution}
Our current evaluation is conducted on $64 \times 64$ resolutions. While modern generative architectures scale readily to higher resolutions (e.g., $128^2$), we restricted our study to this scale to rigorously evaluate physical identifiability under extreme sparsity. 
Two factors motivated this choice: 
\begin{enumerate}
    \item \textbf{Prevention of Hallucination:} As discussed in Section 2.3, sparse observations ($|\omega|=12$) provide limited mutual information. At higher resolutions, high-frequency details become unidentifiable from data alone. Generating them would result in ``posterior collapse'' to the prior, yielding visually plausible but physically deceptive fine structures (artifacts not supported by sensors).
    \item \textbf{Focus on Uncertainty Quantification:} Reliable decision-making requires accurate risk assessment. The moderate resolution allows us to allocate the computational budget to extensive ensemble sampling ($N=200$) for rigorous Uncertainty Quantification (Section 4.4), which we prioritize over perceptual super-resolution in safety-critical domains.
\end{enumerate}
Future work will explore the extension of PIS to high-resolution 3D physical fields or complex domains with non-trivial boundaries, potentially utilizing hierarchical set encoders to manage the quadratic complexity of attention mechanisms.

\subsection{Physical Constraint Integration}
\label{subsec:lim_physics}

PIS currently operates as a probabilistic solver that learns physical priors implicitly from large-scale datasets. While incorporating explicit PDE residuals (e.g., via guidance) could theoretically enhance out-of-distribution reliability, we deliberately prioritize inference efficiency. Explicitly enforcing conservation laws during sampling necessitates computing high-order derivatives at every integration step, which introduces prohibitive computational overhead and undermines the real-time capabilities of the solver. Future work aims to explore lightweight constraint mechanisms that balance strict physical compliance with computational efficiency.

\subsection{Noise Models Beyond Gaussian}
\label{subsec:lim_noise}

In our robustness experiments, we primarily focused on Gaussian measurement noise. Real-world sensors may exhibit non-Gaussian noise profiles, such as outliers, systematic bias, or multi-modal signal corruption. While the generative nature of PIS and its uncertainty quantification (UQ) modules are designed to handle ambiguity, specialized noise-modeling layers or robust likelihood formulations could be integrated to handle these complex error distributions.

\subsection{Temporal Dynamics}
\label{subsec:lim_temporal}

The current framework is optimized for steady-state or fixed-time-step inversion. Extending PIS to time-varying physical processes governed by parabolic or hyperbolic PDEs would require integrating temporal conditioning (e.g., via Recurrent Neural Networks or Temporal Attention) to capture the causal evolution of the physical parameter.


\end{document}